\documentclass[12pt,a4paper]{article}

\usepackage[margin=1in]{geometry}

\usepackage[numbers,sort&compress]{natbib}

\usepackage{amsmath}
\usepackage{amssymb}
\usepackage{bm}

\usepackage{xcolor}
\usepackage{graphicx}
\usepackage[export]{adjustbox}
\usepackage{array}     
\usepackage{makecell}  
\usepackage{booktabs}  
\usepackage[section]{placeins}  

\usepackage[colorlinks,citecolor=blue,urlcolor=blue,linkcolor=blue]{hyperref}

\newlength{\figwidth}
\begin{document}

\begin{center}
\rule{\textwidth}{4pt}\\[0.5em]
{\fontsize{17}{20}\selectfont\bfseries Bayesian Inference of Psychometric Variables\\
From Brain and Behavior in Implicit Association Tests\par}
\vspace{0.3em}
\rule{\textwidth}{1pt}
\vspace{0.8em}
{\fontsize{12}{16}\selectfont\par
\textbf{Christian~A.~Kothe}\textsuperscript{1}\quad
\textbf{Sean~Mullen}\textsuperscript{1}\quad
\textbf{Michael~V.~Bronstein}\textsuperscript{2,4}\\
\textbf{Grant~Hanada}\textsuperscript{1}\quad
\textbf{Marcelo~Cicconet}\textsuperscript{5}\quad
\textbf{Aaron~N.~McInnes}\textsuperscript{2}\quad
\textbf{Tim~Mullen}\textsuperscript{1}\\
\textbf{Marc~Aafjes}\textsuperscript{5}\quad
\textbf{Scott~R.~Sponheim}\textsuperscript{2,3}\quad
\textbf{Alik~S.~Widge}\textsuperscript{2}%
\par}
\vspace{0.5em}
{\footnotesize
\textsuperscript{1}Intheon, La Jolla, CA, USA\quad
\textsuperscript{2}Department of Psychiatry and Behavioral Sciences, University of Minnesota, MN, USA\\
\textsuperscript{3}Minneapolis VA Medical Center, MN, USA\quad
\textsuperscript{4}Institute for Health Informatics, University of Minnesota, MN, USA\\
\textsuperscript{5}Deliberate AI, New York, NY, USA\\[4pt]
\texttt{\{christian.kothe, sean.mullen\}@intheon.io}%
\par}
\end{center}

\vspace{0.5em}

\begin{abstract}
\textit{Objective}. We establish a principled method for inferring mental health related psychometric variables from neural and behavioral data using the Implicit Association Test (IAT) as the data generation engine, aiming to overcome the limited predictive performance (typically under 0.7 area under the ROC curve / AUC) of the gold-standard D-score method, which relies solely on reaction times.

\textit{Approach}. We propose a sparse hierarchical Bayesian model that leverages multi-modal data to predict experiences related to mental illness symptoms in new participants. The model is a multivariate generalization of the D-score with trainable parameters, engineered for parameter efficiency in the high-dimensional, small-cohort regime typical of IAT studies. Data from two IAT variants were analyzed, including a suicidality-related E-IAT ($n=39$) and a psychosis-related PSY-IAT ($n=34$).

\textit{Main Results}. Our approach overcomes a high inter-individual variability and low within-session effect size in the dataset, reaching AUCs of 0.73 (E-IAT) and 0.76 (PSY-IAT) in the best-performing modality configurations, though corrected 95\% confidence intervals are wide (approximately $\pm 0.18$) and results are marginally significant after false discovery rate (FDR) correction ($q=0.10$). Restricting the E-IAT to major depressive disorder participants further improves AUC to 0.79 $[0.62, 0.97]$ (significant at $q=0.05$). Performance is on par with the best tested reference methods (shrinkage LDA and EEGNet) for each task, even though the latter were hand-adapted to these datasets, while the proposed method was not. Point-estimate accuracy was substantially above near-chance D-scores (0.50--0.53 AUC) in both tasks, while maintaining more consistent performance across both tasks than any single reference method.

\textit{Significance}. Our proposed inference framework shows promise in enhancing IAT-based assessment of experiences related to entrapment and psychosis, and potentially other mental health conditions, although further validation on larger and independent cohorts will be needed to establish clinical utility.
\end{abstract}

\section{Introduction}

In mental healthcare settings, diagnoses and treatments begin with inquiry about patient experiences. Diagnostic labels and treatment course therefore depend on patients' ability to accurately describe their experiences. And yet, myriad factors undermine the fidelity of these descriptions. Because mental illness is highly stigmatized, patients may fear that disclosure will usher in differential treatment from providers. They may also fear loss of autonomy if they disclose risk to themselves or others. Even when patients want to tell providers their experiences, their descriptions can be limited by low levels of insight into symptoms, particularly in conditions---like psychosis---that blur the lines of reality. Given these influences, it is not surprising that interview-based tools for the detection of mental health conditions have poor sensitivities, ranging from no more than 41\% (Suicide Ideation / SI) \cite{mchugh2019association} to around 50\% (Major Depressive Disorder / MDD) \cite{carey2014accuracy} across many conditions.

With this in mind, researchers have sought to develop alternative tools for mental health assessment that do not rely on self-reports. A significant portion of these efforts have centered on Implicit Association Tests \cite{greenwald1998measuring} (IATs), building on their widespread use in studies of social and cognitive psychology \cite{mandelbaum2016attitude}. IATs serve to infer subconscious associations among two concepts (e.g.\ Pepsi and Coke) and two attributes (e.g.\ pleasant and unpleasant) \cite{nosek2008associations}. In the traditional IAT, all four categories are mapped onto two responses, but the brief version of the task, known as the Brief-IAT (BIAT) \cite{sriram2009brief} reduces the degree of counter-balancing and instructs subjects to focus on just two category--response mappings. This simplified task is both shorter (i.e., more feasible to administer in healthcare settings) and more accessible to clinical populations with potential cognitive deficits.

The trial-oriented nature and binary classification goal in the (B)IAT setup strongly suggest that this task may be amenable to a multivariate machine-learning (ML) based ``decoding'' approach (e.g., \cite{blankertz2011single}), where activity from simultaneously recorded brain and/or other signals such as EEG could be extracted relative to events and serve as predictors used to infer the psychometric variable of interest. This machine-learning angle is at the heart of our approach, but, as will be discussed, several factors conspire to render the inference task considerably more complex and nuanced than the aforementioned conventional ML setup.

\subsection{Related Work} \label{sec:related}

The consensus method for inference based on IATs is the D-score method (Difference score), which amounts to subtracting the average response latency in one of the paired concept/attribute blocks (of trials) from the average response latency in the other paired concept/attribute blocks, and dividing this difference by the standard deviation of response latency across all trials. As a result, the magnitude of D-scores is understood as an effect size and the sign encodes its directionality (association between a certain pairing of concepts/attributes of interest present vs. absent).

The primary source from which this method draws its effectiveness is the careful balancing of two task conditions, which relies on matched stimuli, block interleaving, and preprocessing of the reaction-time variable. Although this recipe has been sufficient to find relations between implicit cognition and numerous outcomes (e.g., \cite{werntz2016characterizing}, \cite{price2021repeated}), it does not fully mitigate several important shortcomings. D-scores operate on a single IAT readout (reaction time), leaving the method fundamentally limited in its explanatory power and prone to numerous confounds affecting reaction times. D-scores are not naturally formulated to work with other types of features (e.g., brain activity), nor does the approach readily suggest a method for learning feature weightings from data. For these and other reasons, the accuracy of models classifying people based on IAT D-scores remains low (depending on the type of IAT, cohort, data quality, and ground-truthing strategy). As a notable example, D-scores were found to yield AUCs of 0.59--0.67 when predicting suicide attempts in pediatric ER patients \cite{brent2023comparison}. This is not sufficient for clinical utility. The shortcomings of traditional D-scores likely also contribute to their inability to explain practically significant amounts of variation in suicidality after accounting for explicit attitudes \cite{freichel2024explicit}, rendering it unclear whether IAT administration is worthwhile under this analytic regime.

While looking beyond reaction times might enable more information to be extracted from IATs, surprisingly few studies have applied machine learning to multi-modal IAT data with the goal of predicting the primary quantity of interest, that is, implicit biases. One study by Nikseresht et al.\cite{nikseresht2021detection} aimed to predict the participant's implicit bias as measured by the Race IAT, using 91 handcrafted features derived from physiological signals collected with the Empatica E4 wristband, including electrodermal activity (EDA), Photoplethysmography (PPG), heart rate and 3-axis accelerometer data, and using an XGBoost \cite{chen2016xgboost} ML approach; this study achieved an accuracy of 76.1\% in classifying participants as biased or unbiased. However, due to the highly Empatica E4 tailored feature extraction and absent detail on how the XGBoost method was configured, this approach was not readily adaptable to our data modalities and could not be used as a comparison baseline.

A study by \cite{boldt2018detecting} employed facial action units (FAUs) in the context of an IAT task, although their goal differed from ours in that they attempted to decide whether the participant was ``faking'' their IAT results or not. They compared a number of ML methods including Naive Bayes, SVM, and Random Forests, and achieved accuracies in the 70--80\% range; however, their method is not directly transferable to our setting since decoding of the primary IAT variable requires structural matching to the task design (decoding one variable from a contrast between two others), as will be discussed below. Beyond these two studies, we were able to find various uses of multivariate pattern analysis on IAT tasks for diverse exploratory purposes (for example \cite{cala2023eye}, who analyzed systematic differences in gaze patterns using recurrence quantification analysis), but none of these were used in an out-of-sample prediction capacity or attempted to infer the primary task variable.

Thus, for the purpose of comparison we primarily draw on machine learning methods that have proven adaptable to a wide range of trial-oriented cognitive tasks and have a track record of good performance on the modalities that we collected, i.e., EEG, eye tracking, and to a lesser extent facial action units, and we adapt these to our IAT scenario; this is described in more detail in section \ref{sec:refmethods}.

\subsection{The Present Study}

Given the promise of multi-modal ML approaches and the inherent limitations of D-scores, we propose a principled and adaptable method for inference of binary psychometric variables from IAT data, where it is assumed that auxiliary neural, bio and/or behavioral data (such as EEG, eye tracking, or others) are collected while the participant is performing the task. Our method retains the same assumption that underlies D-scores---that differences in responses to trials across IAT block types are informative, instantiated within a modern ML framework and made parameter efficient using structured sparsity.

The present study examines the performance of the proposed method across two use-cases---inference about mental health diagnoses and inference about the presence/absence of a type of clinically meaningful experiences, comparing it to the performance of both existing ML approaches and the traditional D-score method. We show that our method is highly competitive with alternatives and overcomes challenges including small-to-medium training sets and generalization to new participants. It therefore represents an early but encouraging step towards elevating IATs from a research tool to a more informative assessment, by demonstrating that substantially more signal can be extracted from IAT sessions than the prevailing D-score method captures.

\section{Materials and Methods}

\subsection{Task Selection}

Data from two IATs---the Psychosis IAT (PSY-IAT a.k.a. P-IAT) \cite{kirschenbaum2022validation} and the novel Entrapment IAT (E-IAT)---were used in the present study. The PSY-IAT presents participants with words describing psychosis-relevant experiences (e.g., ``visions'') in order to probe implicit associations between the self and psychosis. The E-IAT is a novel task developed to examine implicit attitudes regarding entrapment, which is thought to be a central driver of depression and suicidality \cite{gilbert1998role,o2018integrated}. In the present study, the E-IAT serves primarily to explore the generality of our method across a second, structurally identical but thematically distinct IAT, operationalized against a different type of dependent variable (self-report derived from ecological momentary assessment (EMA) rather than clinical diagnosis).

\subsection{Task Design}

\begin{figure}[htbp]
\centering

\begin{minipage}{\figwidth}
  \raggedright
  \hspace{1pt}($a$)\vspace{-2pt}\\
  \includegraphics[width=\figwidth]{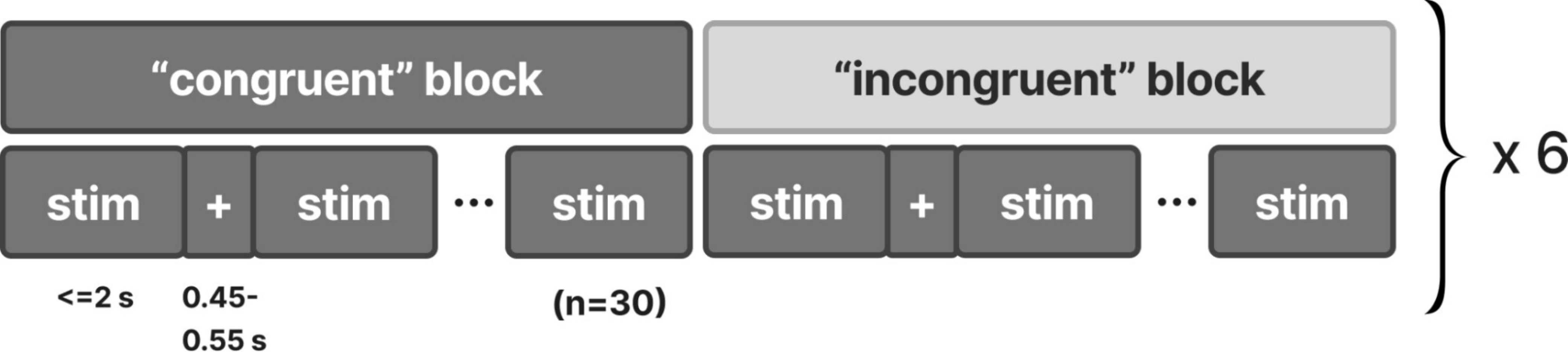}
\end{minipage}

\vspace{6pt}

\begin{minipage}{\figwidth}
\begin{minipage}[t]{0.48\linewidth}
  \raggedright
  \hspace{1pt}($b$)\vspace{-2pt}\\
  \includegraphics[width=\linewidth]{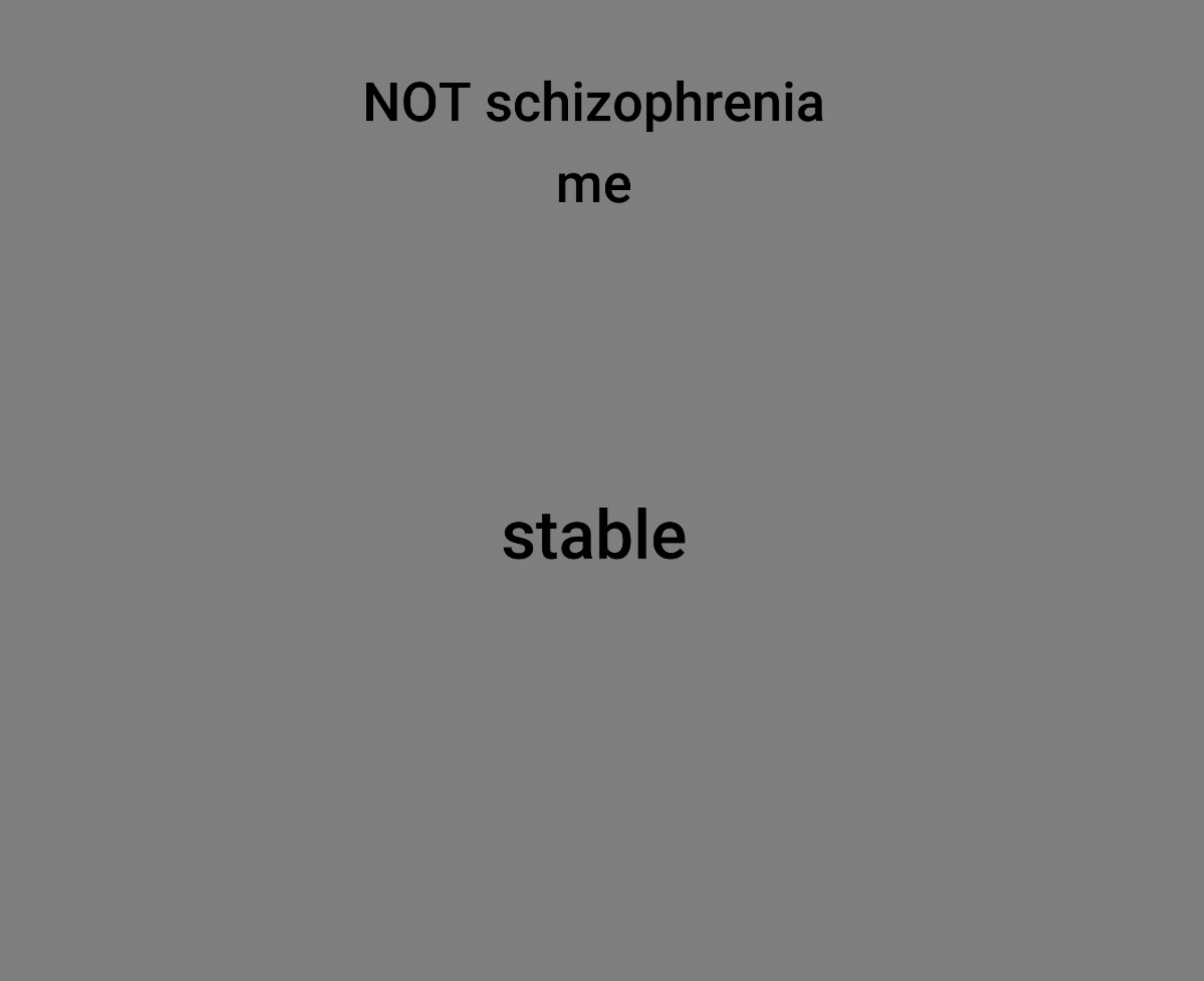}
\end{minipage}%
\hfill
\begin{minipage}[t]{0.48\linewidth}
  \raggedright
  \hspace{1pt}($c$)\vspace{-2pt}\\
  \includegraphics[width=\linewidth]{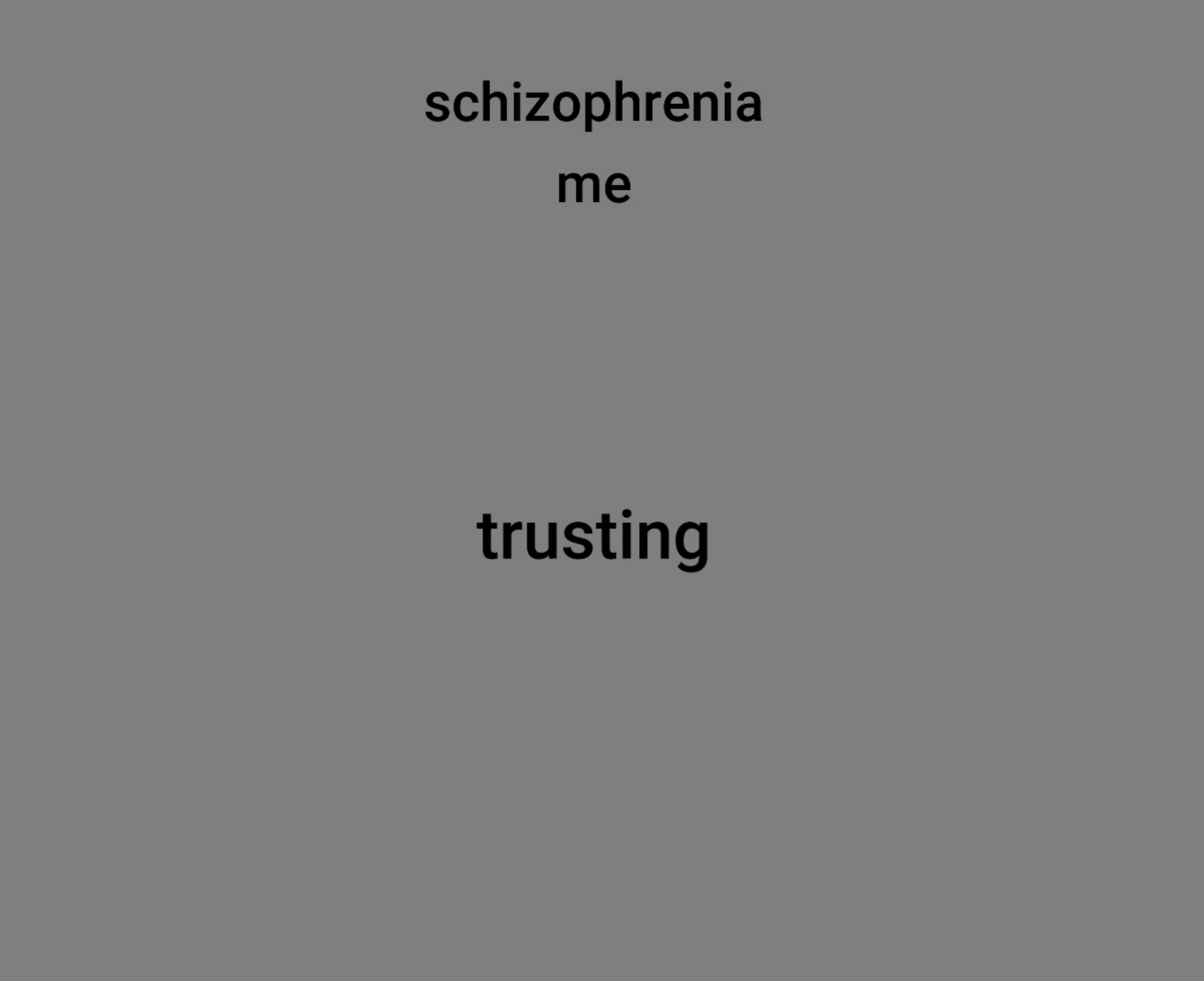}
\end{minipage}
\end{minipage}
\caption{
($a$) Block/trial structure used for E-IAT and PSY-IAT task designs. A ``congruent'' or ``incongruent'' block represents a block of trials where the concept-attribute pair (i.e., ``trapped/me'') displayed is in agreement or conflict, respectively, with the subject type (patient / control). ($b$,\,$c$) Screenshots of PSY-IAT task showing a trial from the two block types. ($b$) depicts an incongruent trial for a person with schizophrenia (a congruent trial for a person without schizophrenia). ($c$) depicts a congruent trial for a person with schizophrenia (incongruent for a person without schizophrenia).
}
\label{fig:task-design}
\end{figure}

The structure of both tasks was identical (i.e., only the task content differed) and was based on the BIAT (cf. \cite{sriram2009brief} for a good introduction). We note however that our method is also applicable to full-length IATs by ignoring the additional experimental conditions present there, but the method was not tested with a full-length IAT. Two modifications to the BIAT's structure were made for the present study: the concept-attribute labels were displayed in the top center of the screen and aligned vertically (instead of the concept displayed top-left and the attribute displayed top-right) as shown in figure~\ref{fig:task-design}, to minimize left-right eye saccades that might factor into the physiological measures being analyzed. We also lengthened the task beyond the original 6 blocks of 20 trials each, with the goal of providing our model with more data allowing it to better reproduce the high-dimensional variables involved. For the E-IAT, we collected 12 blocks of 30 trials each, for a total of 360 trials. For the PSY-IAT, we collected 12 blocks of 28 trials, for a total of 336 trials. Another example of the BIAT template is the Death Brief Implicit Association Task (D-BIAT) \cite{millner2018brief}, which we also administered, in a similarly modified fashion. For full details on our modified task design and the stimulus content, please see Appendix~\ref{app:taskdetails}.

Before attempting to classify participants based on IAT results, we generally discard data from trials showing stimuli that belonged in the ``not-me'' category (of which there were 5 unique stimuli, i.e., ``they'', ``them'', etc.). These stimuli are discarded because the correct response would always be ``not associated,'' rendering these less informative. As a result, the number of trials, for each session, seen by the model, was 270 for the E-IAT and 252 for the PSY-IAT.

\subsection{Physical Setup and Sensor Suite}

Sensor data were collected while subjects performed a given task (E-IAT or PSY-IAT) in a dedicated data collection room. EEG was captured using a BioSemi ActiveTwo system and using the vendor's 128-electrode ABC layout, downsampled from 2048 Hz to 512 Hz at acquisition time, while video was recorded using a FLIR Blackfly S RGB camera (running at 60 Hz). Pupillometry and eye tracking data were collected using a Tobii Pro Nano infrared eye tracker mounted along the bottom of the monitor, also at 60 Hz. Multi-modal data capture and time synchronization was performed using the Lab Streaming Layer \cite{kothe2025lab}.

\begin{figure}[htbp]
\centering
\begin{minipage}[t]{0.38\textwidth}
  \raggedright
  ($a$)\\[2pt]
  \includegraphics[width=\linewidth]{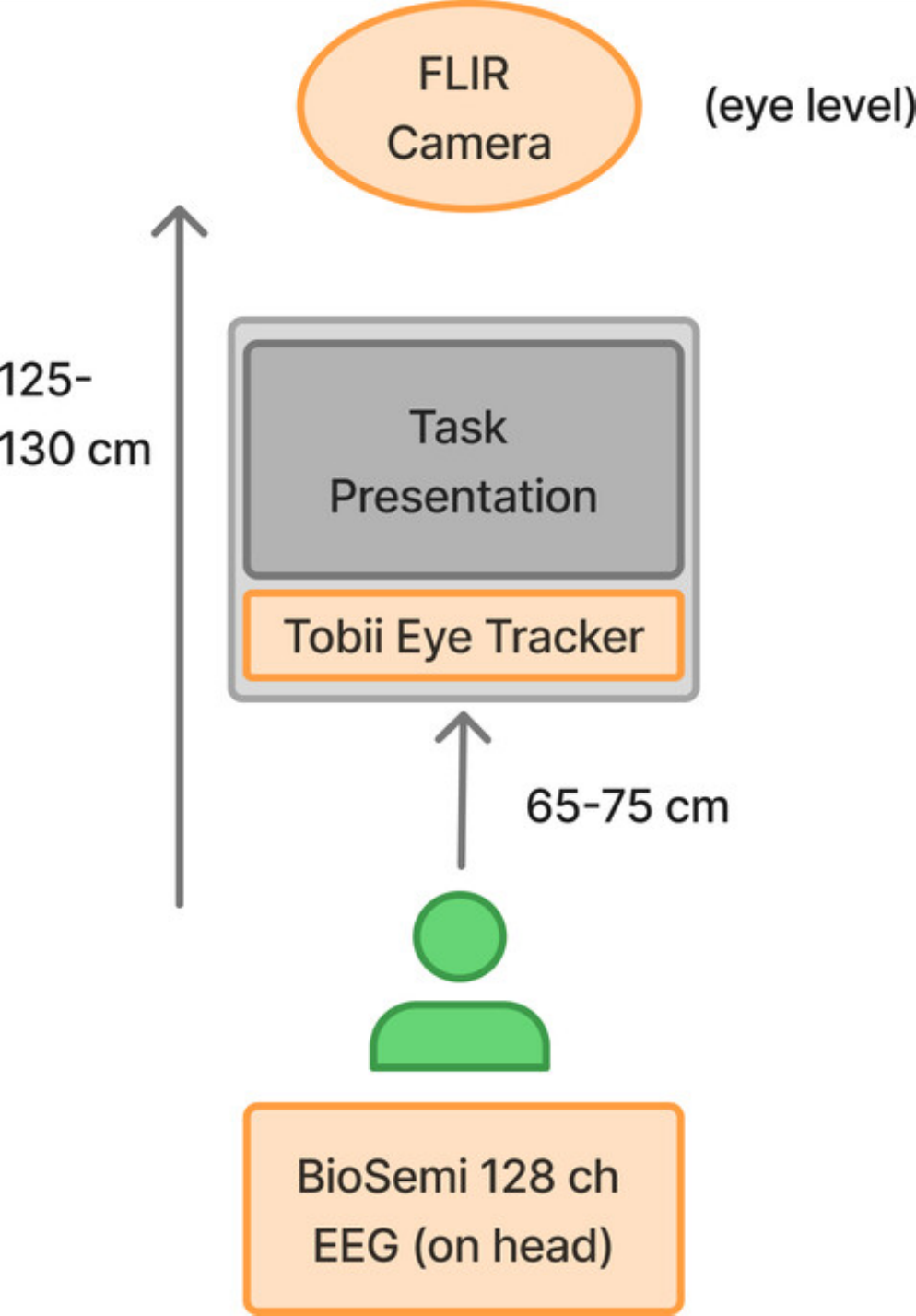}
\end{minipage}%
\hfill
\begin{minipage}[t]{0.58\textwidth}
  \raggedright
  ($b$)\\[2pt]
  \includegraphics[width=\linewidth]{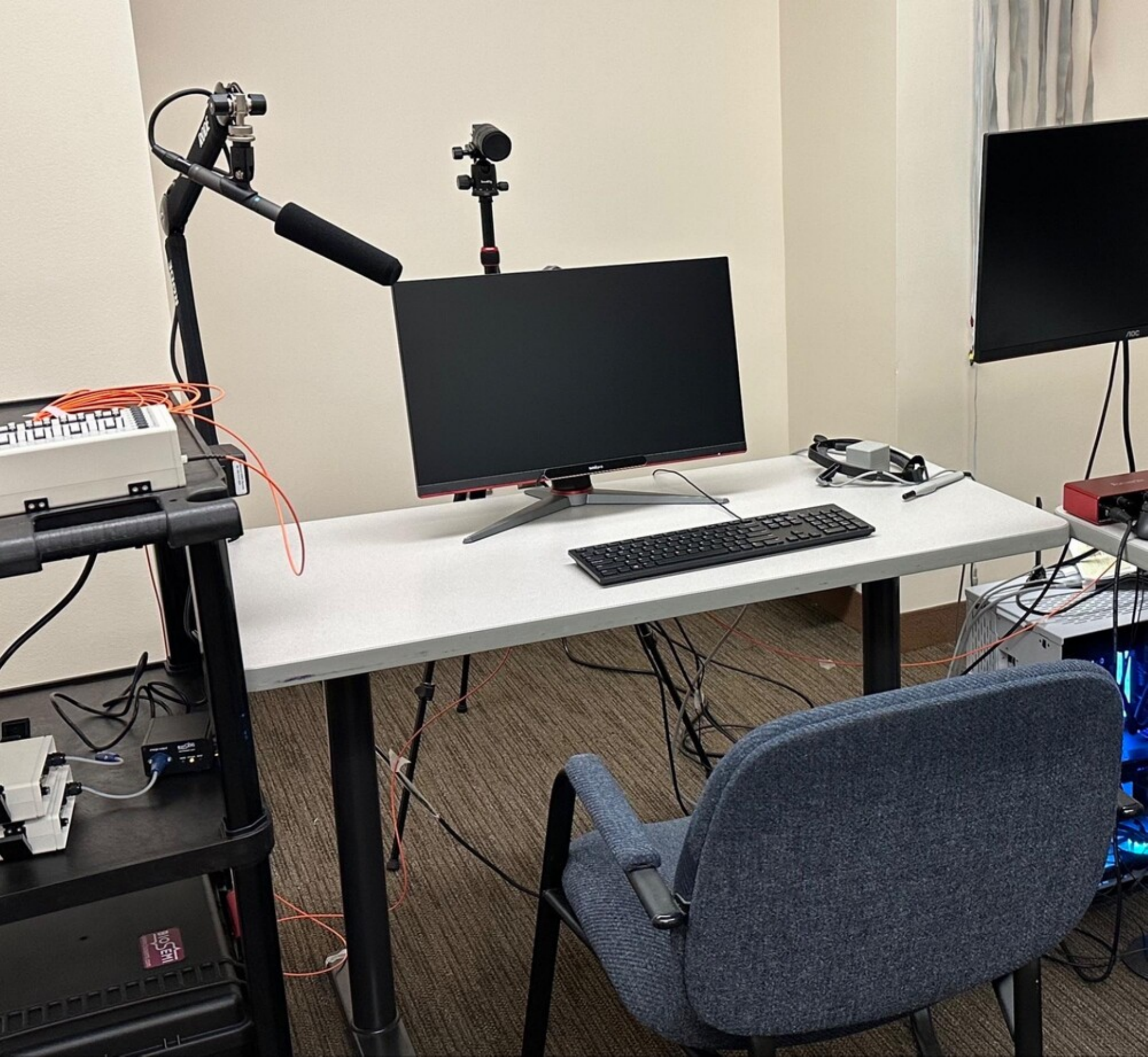}
\end{minipage}
\caption{
($a$) A diagram of the data capture setup showing the positioning of the sensors relative to the subject. ($b$) A photograph of the data collection setup at the University of Minnesota used for the E-IAT task data collection. The same setup was used for the collection of the PSY-IAT task at the Minneapolis VA Medical Center.
}
\label{fig:setup}
\end{figure}

The task was presented on a desktop PC using Intheon's (La Jolla, CA) Experiment Recorder task presentation software and NeuroPype\footnote{\texttt{https://neuropype.io}} software for signal data capture, running on Windows 11. Each task was collected as part of a larger task battery that included other tasks and questionnaires that were not evaluated here (the order of the tasks was balanced between subjects). Camera data collection was implemented using software developed and tailored to this study by Deliberate AI (New York, NY). A schematic and photograph of the setup are shown in figure~\ref{fig:setup}.

\subsection{Populations and Datasets}

This study was performed in accordance with the Declaration of Helsinki. This adult human subject study was approved by Advarra IRB; ethical review was ceded to this board by the University of Minnesota IRB. All participants provided written informed consent to participate in this study.

Data collection was conducted as part of a larger multi-site study on suicide ideation led by the University of Minnesota. The PSY-IAT and E-IAT were secondary tasks in this study. The PSY-IAT was collected at the Minneapolis VA Medical Center, with the experimental group consisting of adults with a current or past history of non-pharmacologically induced psychosis, and a control group drawn from the general population ages 18--65 with no diagnosis of a mental illness.

The E-IAT was collected at the University of Minnesota Department of Psychiatry, with the experimental group consisting of adults receiving treatment for major depressive disorder (MDD), and a control group (CTL) drawn from the general population ages 18--65 and self-reportedly without any diagnosis of a mental illness. MDD participants in the E-IAT sample were recruited from local interventional psychiatry clinics, and were required to be on waitlist to receive transcranial magnetic stimulation or ketamine therapy.

The larger study was focused on suicide ideation with Active SI as the primary metric and recruitment was not directly based on characteristics central to the E-IAT (such as levels of perceived entrapment) since it was a secondary task. All participants were also required to own a smartphone compatible with the EMA application used to ascertain E-IAT ground truth, or willing to use a study loaner phone. Exclusion criteria were contraindications to EEG (e.g., visible scalp abrasions), pregnancy, recent psychiatric hospitalization, or active military or Department of Defense employment, due to sponsor requirements.

The order of the two IAT tasks (PSY-IAT, D-BIAT in the VA cohort, or E-IAT, D-BIAT at UMN) was balanced across subjects in each group.

The sample consisted of all eligible participants who were recruitable at each site during a fixed funding period (April 2023 to September 2025). No formal power analysis was performed to determine sample size. A post-hoc sensitivity analysis using the Nadeau \& Bengio \cite{nadeau1999ttest} corrected variance framework indicates that, at our sample sizes and observed fold-level variability, the minimum detectable effect size (MDES) at 80\% power and $\alpha=0.05$ corresponds to a true AUC of approximately 0.70 (E-IAT) and 0.72 (PSY-IAT) for the respective best-performing configurations. This is offered as a characterization of the study's detection limits, not as evidence of adequate power (cf.\ \cite{hoenig2001abuse}). Our observed point-estimate AUCs of 0.73 and 0.76 exceed these thresholds, but the corrected 95\% confidence intervals (e.g., [0.56, 0.91] for the main full-sample E-IAT configuration) are wide, reflecting the modest sample sizes, and readers should interpret reported AUC values with this uncertainty in mind.

Additional exclusion criteria for analysis were (1) incomplete session record across the relevant data being analyzed (EEG, eye tracker, video, event markers, questionnaire responses) due to data collection issues, and (2) behavior indicating improper performance of the task due to lack of task understanding, failure to be awake and alert during the task (per experimenter notes), or abnormally high error rates during task performance. The PSY-IAT was performed by 44 subjects, of whom 7 were excluded due incomplete/missing video data. Another three were excluded due to task behavior. Of the 34 subjects analyzed, 17 were from the experimental (psychosis) group, and 17 from the control group. The E-IAT was performed by 43 subjects, of which two were excluded due to incomplete data, and two were excluded due to task behavior. Of the 39 subjects included in the analysis, 33 were from in the MDD group and 6 in the control group. Note that the subject populations for the two tasks were from two separate sites and were non-overlapping.

\subsection{Ground Truthing (Dependent Variables)}

In the case of the PSY-IAT, the ground truth for subject labels was the clinically documented presence or absence of a history of psychosis. This was also the criterion for the experimental and control group assignment and was used as the dependent variable for the classification reported here.

Ground truth for the E-IAT was generated from self-reports. These self-reports were collected through ecological momentary assessments (EMAs), surveys completed by participants via their smartphones as they go about their daily lives. Participants were asked to complete these surveys 3x/day for seven days prior to IAT data collection. The EMA battery generally covered constructs relevant to suicide (e.g., psychological pain, hopelessness). The six items specifically related to entrapment (and defeat, which we consider the same construct) were adapted from a widely used self-report scale \cite{griffiths2015development} and were internally consistent (within-persons omega-total = .73, between-persons omega-total = .97). Item ratings were summed together at each EMA timepoint, then averaged across timepoints to obtain a total entrapment score for each participant. The entrapment score distribution was approximately uniform (excess kurtosis $= -1.24$, KS test vs.\ uniform $p = 0.50$; Hartigan's dip test for bimodality $p = 0.80$; similar for the MDD-only subset: kurtosis $= -1.14$, KS $p = 0.80$, dip $p = 0.78$), with near-zero skewness. This score was binarized at the whole-sample median (1.87), splitting participants into ``high'' (above-median) and ``low'' (below-median) levels of entrapment so as to conform to the overall binary detection framework investigated in this article. This binarized version was used as our dependent variable. Splits were not done separately by participant group because participants were not recruited based on entrapment levels, but rather on characteristics such as mental health diagnosis histories. All control-group (CTL) subjects in the E-IAT dataset fell into the low-entrapment group and among the MDD subjects, 12 (36\%) were low-entrapment and 21 (64\%) were high-entrapment. To rule out that the ML model is partly acting on a CTL/MDD contrast instead of, or in addition to, the entrapment contrast (and is therefore confounded), we also include a side analysis of ML performance on the MDD group only; that analysis shows that our model performs considerably better on MDD-only than on the broader MDD+CTL population, however we report here primarily results for the full sample since that was our pre-specified analysis and would also be more reflective the (presumed heterogeneous) broader population.

\subsection{Design and Testing Philosophy}

The method described in this article is, by its multi-modal nature, inherently complex and has a number of preprocessing and modeling parameters. We refrain from the common practice of adapting these parameters in a data-contingent manner, perhaps so as to improve performance, or in making apparently reasonable choices after having reviewed the collected dataset via statistical summaries or visually. Rather, given the relatively modest size of datasets involved here, we judge it to be virtually impossible to do so without inherently overfitting the approach to the dataset at hand in myriad ways, as called out in \cite{gelman2013garden}. Avoiding this required relying on a largely pre-specified model and applying this model essentially unchanged to new data. In line with this, the approach described here was originally developed on a series of separate tasks and datasets, including variants of a custom autobiographical recall task and the D-BIAT. In this article we describe both the model and its application to two \textit{new} tasks to which it had not previously been applied, essentially serving as a first description and validation study of the model; a separate article discusses the application of the model described herein to one of these prior tasks (the D-BIAT).

The transfer to new datasets is more likely to succeed if the model is largely free of parameters that are sensitive to a particular dataset and would require retuning. To make this possible we made the following design choices in our approach:
\begin{itemize}
  \item Relying primarily on extensively ``battle-''tested generic preprocessing components for the various modalities, avoiding dataset-specific preprocessing, and using only default or otherwise conventional parameter choices for data preparation and preprocessing.
  \item Avoiding potentially task-specific choices in the model (e.g., data-informed feature extraction) entirely and using generic structural prior knowledge instead.
  \item Relying primarily on statistical modeling that automatically adapts to the complexity of the data at hand, using hierarchical Bayesian techniques.
  \item Choosing only weakly informative ``default'' configurations for any remaining components.
  \item For parameters that are suspected to be somewhat dataset specific we reproduce multiple settings in the results side by side.
\end{itemize}

When comparing with reference methods from the literature (see section~\ref{sec:refmethods} for an overview and Appendix~\ref{app:refmethods} for full details), a challenge was that there were no existing methods that were reasonably well adapted to our combination of IAT task and data modalities. We addressed this by using generic methods or methods previously tested on similar data, which left the question of how to set their parameters (for example feature extraction) in a way that does not unfairly penalize these methods. To account for this, we tailored their parameters somewhat to the two tasks at hand, in some cases presented multiple choices side by side, and we show that the proposed method remains competitive with and/or outperforms these reference methods even in a setting that is favorable to the latter.

\subsection{Preprocessing} \label{sec:preproc}

To prepare data for use with the predictive model, we separately preprocessed each recording in a modality-specific manner using a previously developed robust preprocessing pipeline. For EEG, we employed the data cleaning pipeline described in \cite{mullen2015real} as implemented in the NeuroPype software (Intheon, La Jolla, CA). In the following we briefly recap the main EEG preprocessing steps and their parameters; for additional details, see the associated references or the open-source implementation in the \texttt{clean\_rawdata}\footnote{\texttt{https://github.com/sccn/clean\_rawdata}} plugin for EEGLAB \cite{delorme2004eeglab}.

First, we assigned 3D sensor/channel locations in a head-relative coordinate system according to BioSemi's cap-specific coordinates file; session-specific coordinate measurements were not used. We then removed any non-EEG channels, resampled the time series to 128 Hz using a polyphase implementation (\texttt{resample\_poly} function provided by SciPy \cite{virtanen2020scipy}), removed the DC offset by subtracting the per-channel median, and then applied a zero-phase (forward-backward) FIR highpass filter with a transition bandwidth of 0.25--0.5 Hz and 120 dB stop-band suppression. We subsequently rejected noisy channels whose high-frequency (above 45 Hz) content (magnitude in robust z-scores relative to all channels in the recording) exceeded 4 st.\ dev.\ or whose correlation with a robust estimate of the channel was below 0.7. The robust estimation (RANSAC) was performed as in \cite{bigdely2015prep} and used 200 pseudorandom subsets of 15\% of channels each. We then projected high-amplitude artifacts out of the data using the artifact subspace reconstruction (ASR) method \cite{mullen2015real} using a cutoff of 15 st.\ dev., and subsequently rejected any time windows that contained residual high-amplitude artifacts if their power in at least 25\% of channels exceeded a $-4$ to $+7$ st.\ dev.\ range (also using robust z-scores across all channels). Next we suppressed line noise and other high-frequency content using a zero-phase FIR lowpass filter with 40--45 Hz transition band and 50 dB stopband suppression, and interpolated any previously rejected channels using spherical spline interpolation \cite{perrin1989spherical}. Lastly, we re-referenced all channels to the common average and retained a subset of 64 equidistant channels using a method analogous to that of the METH toolbox\footnote{Guido Nolte's MEG \& EEG Toolbox of Hamburg}. A reduction from 128 to 64 channels had previously been identified as a good tradeoff of model degrees of freedom vs.\ spatial resolution in analyses of the D-BIAT and earlier autobiographical tasks and was retained for this study in line with our parameterization philosophy. All other preprocessing parameters match NeuroPype's default processing pipeline for machine learning, except for a more forgiving ASR cutoff (default is 10) and a highpass filter better adapted to event-related potentials.

From the participant videos, we extracted Facial Action Units (FAUs). FAU measurements employed a preprocessing pipeline by Deliberate AI (New York, NY), which relies on MediaPipe \cite{lugaresi2019mediapipe}, OpenFace \cite{baltruvsaitis2015cross}, and custom libraries to derive time-varying FAU estimates according to the Facial Action Coding System (FACS) \cite{ekman1978facial} plus head orientation and gaze vectors, along with the discrete first and second time derivatives of all channels (i.e.\ ``dynamic'' features). The following FACS-defined AUs were extracted by this pipeline: 1, 2, 4, 5, 6, 7, 9, 10, 12, 14, 15, 17, 20, 23, 25, 26, and 45.
For the eye tracking modality we use all channels delivered by the Tobii eye tracker, including estimated screen coordinates, pupil diameter, and pupil confidence. We also corrected (delayed) stimulus event time stamps according to a photodiode-measured end-to-end on-screen stimulus presentation delay of 54 ms. We use the following acronyms for these modalities in the remainder of the article: \textit{EEG} (preprocessed EEG), \textit{gaze} (eye tracker), \textit{FAU} (facial action unit estimates), and \textit{Dyn} (various dynamic / time-derivative features).

Following continuous-data preprocessing, stimulus-locked segments were extracted from each modality for each trial. We rejected segments where time windows that had previously been removed due to artifacts fell within a range of $-0.5$ to $1.0$ seconds relative to stimulus onset, and extracted the remaining segments spanning this same time range from FAU and gaze streams, respectively. For EEG, to reduce the degrees of freedom of the model, we retained shorter time segments spanning $-0.1$ to $0.45$ seconds; this time window was previously determined on the D-BIAT task. This yielded, for a given trial $t$, an EEG segment $\bm{X}^{\mathrm{EEG}}_{t} \in \mathbb{R}^{C^{\mathrm{EEG}} \times K^{\mathrm{EEG}}}$ where $C^{\mathrm{EEG}}=64$ is the number of EEG channels and $K^{\mathrm{EEG}}=71$ is the number of EEG time points, along with an FAU segment $\bm{X}^{\mathrm{FAU}}_{t} \in \mathbb{R}^{41 \times 96}$, a Dyn segment $\bm{X}^{\mathrm{Dyn}}_{t} \in \mathbb{R}^{41 \times 96}$, and a gaze segment $\bm{X}^{\mathrm{gaze}}_{t} \in \mathbb{R}^{6 \times 96}$, of which not necessarily all modalities will be used in a given model. To keep notation light, we omit the trial and modality sub/superscripts from symbols where clear from the context.

Before training our predictive models, we first standardized each (zero-mean) channel in each modality $m$ to unit variance on the respective training set. In line with standard practice we treat the vector of scales $\bm{s}_m \in \mathbb{R}^{C^m}$ as a learnable model parameter that is re-fit on each training set and applied (but not recomputed) on the respective test set(s).

To prepare data for traditional neural analyses, specifically the event-related potential (ERP) figures (section~\ref{sec:neural}), the preprocessing steps were largely similar to what is described above with minor modifications in line with more typical processing chains used in ERP studies. We set the FIR highpass filter with a transition bandwidth of 0.1--0.2 Hz, the FIR lowpass filter with a transition bandwidth of 15--20 Hz, and set ASR cutoff threshold to 10 st.\ dev.. ERPs were furthermore baseline-corrected to 100 ms of pre-stimulus activity after segment extraction. For group-level statistics, we use mass univariate ANOVAs on various factors (e.g.\ block conditions, clinical status, etc.) for both within-subjects and between-subjects conditions.

\subsection{Predictive Model} \label{sec:model}

\subsubsection{Goal} \label{sec:goal}
Our model is designed to accept a recording (``session'') worth of (appropriately segmented) brain and/or behavioral time-series data collected from a person across one or more measurement modalities while performing the IAT. It then predicts from these measures the probability that the person displays a positive implicit association with the psychometric variable being tested. This also yields (through subtraction from 1) the probability of the negative event, that is, the person \textit{not} displaying a positive association. For simplicity, we equate sessions with participants in the following, i.e., one participant yields one session, as this is the structure of both our datasets.

\subsubsection{Overall Framework} \label{sec:framework}

Our approach will be to design a type of hierarchical generalized linear model (GLM), optionally multi-modal, that explains the dependent variable $y$ (1 if positive or 0 otherwise) as a function of the given per-session observations, along with model parameters that we will learn from a given training sample. The model therefore has to generalize from the training cohort of participants to a new participant; this is made challenging due to the relatively modest number of participants in typical IAT study datasets, including the ones analyzed here (with $n=$ 39 for the E-IAT and 34 for the PSY-IAT). The relatively large number of explanatory variables in the modalities of interest (EEG, FAU) across the potentially relevant time points and channels leads to many degrees of freedom $df$ in the model (10,000s if not carefully controlled). While this could be reduced with detailed task- and modality-specific feature engineering (e.g., encoding presumed relevant channels and/or time slices), the necessary knowledge is often not available in a new and/or relatively under-studied machine-learning task such as the one studied here, and would have to be inferred from the data at hand alongside the rest of model. As noted, we refrain from this here and we instead employ a (hierarchical) Bayesian approach, which allows us to impose strong structural priors to control model complexity without requiring task-specific knowledge. A key feature of our model is that we rely on a sparse Bayesian learning (SBL) strategy, which achieves high parameter efficiency and thus low effective model complexity, while comparing favorably to the convex $l_1$ penalty (a.k.a. the LASSO) popular in machine learning due to the latter suffering more from over-shrinking of non-zero model coefficients \cite{carvalho2010horseshoe}.

A second key strategy is to fit the model at the (pooled) \textit{single-trial level}, thus leveraging single-trial variability as a stand-in for participant-level variability that we have not observed (for lack of a larger dataset). We will discuss these tradeoffs in the next section. In the following we first present the prediction rule in frequentist terms and then lay out the Bayesian model used to infer the various parameters.

\subsubsection{Prediction Rule} \label{sec:rule}

We aim to estimate $P(y=1 \mid \bm{X},\: \bm{\theta})$ where $\bm{X}$ are the observed/independent variables (measurements), $y$ is the dependent variable, and $\bm{\theta}$ is the collection of model parameters. While the model can in principle accommodate a variety of forms for $y$ (e.g., binary, continuous, or ordinal) using appropriate likelihood functions, we restrict ourselves here to the elementary binary (detection) case, which is well matched to the detection task at hand. We model the session-level success probability as $P(y=1 \mid \bm{X},\: \bm{\theta}) = \mathrm{logistic}(z)$ where $z \in \mathbb{R}$ is the session-level logit and is ultimately a weighted linear combination of appropriately segmented observations from the session of interest, which makes our model a type of logistic regression. The output of the model is interpretable as a detection probability that may be thresholded by the practitioner according to a decision threshold that achieves, for example, a suitable sensitivity/specificity tradeoff. In our setup and for evaluation purposes, we set the decision threshold at the canonical $P(y=1)=0.5$.

The model has to integrate evidence (i.e., logits) across each individual trial $t \in \{1, \ldots, T\}$ of the experimental session to yield $z$. We employ the formula $z = \frac{\omega}{T}\sum_{t=1}^{T} z_t$; that is, $z$ is an average of the per-trial evidence scores $z_t$ weighted by a scale factor $\omega \in \mathbb{R}$. $\omega$ can be interpreted as a ratio of session-level evidence (logit magnitude) to single-trial level evidence and is the sole parameter of the model that is fit at the session level, as part of a two-stage fitting procedure (see section~\ref{sec:confcalib} for more detail).

To streamline notation we treat all data modalities to have generally matrix-variate per-trial observations, so for any given modality $m$ and trial $t$ we observe a matrix $\bm{X}^{m}_{t}$. For a single modality (for example, EEG only), we thus have $z_t = \langle \bm{X}^{m}_{t}, \bm{W}^{m}_t \rangle_{\mathrm{F}}$, i.e., a linear function of the data, where we denote by $\langle \bm{A}, \bm{B} \rangle_{\mathrm{F}} := \mathrm{trace}(\bm{A}^\top \bm{B})$ the Frobenius inner product of two matrices, and where $\bm{W}^{m}_t$ is the learnable weight matrix for modality $m$. In the multi-modal case we treat $z_t$ as a weighted sum across modalities, using a separate per-modality weighting $\alpha_m$. Thus, in the general case of $M$ modalities we have:
\begin{equation} \label{z_t_eq}
z_t = \sum_{m=1}^{M} \alpha_m \langle \bm{X}^{m}_{t}, \bm{W}^{m}_t \rangle_{\mathrm{F}}.
\end{equation}

We write the weight matrices as (technically) trial-dependent, and this is a crucial feature of our model that has its roots in the standard D-score method for analyzing IAT tasks: in D-scoring, $D$ (the analog to our $z$) is the (scaled) mean of all reaction times (observations) in the incongruent condition minus the scaled mean of reaction times for the congruent condition, as in
\[
D = \frac{M_{\mathrm{incongruent}} - M_{\mathrm{congruent}}}{SD_{\mathrm{pooled}}}.
\]
We can capture this in our model by making our weights formally condition-specific, such that, for each \textit{in}congruent trial $t \in \mathcal{I}$ we fix $\bm{W}^m_t$ to a single learnable matrix $\bm{W}^m_{\mathcal{I}}$ and likewise for each \textit{con}gruent trial $t \in \mathcal{C}$ we fix $\bm{W}^m_t=\bm{W}^m_{\mathcal{C}}$. The difference structure in the above formula then suggests a simple ``mirror'' constraint $\bm{W}^m_{\mathcal{I}}=-\bm{W}^m_{\mathcal{C}}$ on the weights. Clearly there exists also a variant that does not impose a sign reversal, but this results in a model that is not a congruency detector but which instead aims to directly decode the participant label, regardless of the block condition. We briefly touch on this in the context of the reference methods (Appendix~\ref{app:refmethods}). Another variant would allow the two weight matrices to be fully \textit{independent}---however, analysis of IAT tasks is extremely sensitive to the balance of the two (symmetric by design) conditions, which likely requires statistical constraints on such a pair of matrices, for example on the norm of the differences; the design space for this is large and we do not explore these options in this article. For the aforementioned balancing reason, our model also includes no intercept parameter, although such a parameter may be included if additional auxiliary predictors (e.g., questionnaire responses, etc.) were added to the model.

Beyond the basic prediction rule outlined above, when the learned posterior distribution over relevant parameters $\bm{\theta}$ is used in the formula, this gives rise to the posterior predictive distribution for $P(y=1 \mid \bm{X}, \bm{\theta})$.

\subsection{Learning} \label{sec:learning}

We pose learning of the relevant model parameters as a Bayesian inference problem where we model the binary outcome variable $y$ using a Bernoulli likelihood given the logits $z$, i.e.,
\begin{equation} \label{eq:likelihood}
y \sim \mathrm{Bernoulli}(\mathrm{logistic}(z)).
\end{equation}
When defining prior distributions for the remaining parameters, our strategy will be to encode primarily \textit{structural} knowledge into the priors. Fundamentally, the type of prior for a weight matrix $\bm{W}^m$ (or by symmetry, its mirror-constrained variant) depends on the nature of the modality and how we expect the effect of interest to manifest in the weights. Due to the mirror constraint, our weight matrix behaves effectively as a trial-level (in)congruency detector and, when viewed as a type of (generalized) linear model, these weights may be understood to a first approximation as modeling a ``contrast'' between observations of congruent vs.\ incongruent IAT trials, respectively. In practice, this being a decoding model, the weights will furthermore involve correction terms that perform suppression and potentially cancellation of nuisance interference or noise.

The simplest type of prior we may consider is the standard normal, $\bm{W}^m \sim \mathcal{N}(0, 1)$, which may be appropriate for modalities with very few predictors (e.g., the scalar per-trial (log-)reaction time). However, already for the 6-channel gaze modality, we have, at 60 Hz sampling rate, well over 500 degrees of freedom in the model---likely too large to allow for reliable fitting from even a 50-participant dataset. We offer a back-of-the-envelope attempt to quantify this in Appendix~\ref{app:samplesize}. Consequently, for this and the more complex time-series observations, a stronger prior is very likely needed. We have found the following assumptions to be sufficient for a range of typical behavioral time-series modalities: optionally smoothness (slow changes) in weights along the time dimension of the weight matrix, and group-wise sparsity along the channels dimension of the matrix. Specifically, we model $\bm{W}^{\mathrm{gaze}}$ and $\bm{W}^{\mathrm{FAU}}$ as both group-sparse and smooth, and $\bm{W}^{\mathrm{Dyn}}$ as group-sparse (but not necessarily smooth). We formalize these assumptions using a grouped horseshoe prior \cite{xu2016bayesian} for the Dyn modality, and a grouped horseshoe where the unscaled weight matrix is drawn from a Gaussian random walk (GRW) process with non-zero initial intercept for the gaze and FAU modalities. Formally, we have

\begin{align}
\tau^{m} &\sim \mathrm{HalfCauchy}(1.0),\; \tau^{m} \in \mathbb{R}_{+}\\
\bm{\lambda}^{m} &\sim \mathrm{HalfCauchy}(1.0),\; \bm{\lambda}^{m} \in \mathbb{R}_{+}^{C^m}\\
\bm{\beta}^{\mathrm{Dyn}} &\sim \mathrm{Normal}(0.0, 1.0),\; \bm{\beta}^{\mathrm{Dyn}} \in \mathbb{R}^{C^{\mathrm{Dyn}} \times K^{\mathrm{Dyn}}}\\
\bm{\beta}^{\mathrm{gaze}} &\sim \mathrm{GRW}(\sigma_0, \sigma^{\mathrm{gaze}}_\mathrm{i}),\; \bm{\beta}^{\mathrm{gaze}} \in \mathbb{R}^{C^{\mathrm{gaze}} \times K^{\mathrm{gaze}}}\\
\bm{\beta}^{\mathrm{FAU}} &\sim \mathrm{GRW}(\sigma_0, \sigma^{\mathrm{FAU}}_\mathrm{i}),\; \bm{\beta}^{\mathrm{FAU}} \in \mathbb{R}^{C^{\mathrm{FAU}} \times K^{\mathrm{FAU}}}
\end{align}

where GRW is the Gaussian random walk prior with initial intercept scale $\sigma_0=1.0$ and learnable innovation scale $\sigma^m_\mathrm{i}$.
Following the grouped horseshoe construction, the resulting sparse weight matrix is then $\bm{W}^m=\tau^{m} \operatorname{diag}(\bm{\lambda}^{m}) \bm{\beta}^{m}$, i.e., a product of a global scale factor $\tau^{m}$, a diagonal matrix of per-channel scales $\operatorname{diag}(\bm{\lambda}^{m})$, and the unscaled weight matrix $\bm{\beta}^{m}$.

For EEG, the situation is complicated considerably by the fact that relevant EEG activity is typically \textit{not} confined to a small set of channels owing to electrical volume conduction, and thus $\bm{W}^{\mathrm{EEG}}$ is generally not group-sparse in channels. However, we may instead assume that only a few latent brain \textit{sources} differ in their activity across the contrast of interest (congruent vs incongruent) so that the weight matrix is effectively group-sparse in terms of brain sources. We choose here a formulation of this that is to our knowledge novel, and which works as follows; this construction constitutes a core part of our method for the EEG case. We model the scalp forward projection of the relevant brain activity contrast as a latent matrix $\bm{A} \in \mathbb{R}^{C \times K}$ whose prior will be chosen to encode the latent source-level sparsity and smoothness, which we choose to be a matrix-normal distribution as in
\begin{equation} \label{eq:A}
\bm{A} \sim \mathrm{MatrixNormal}(\bm{0}, \bm{\Sigma}_{\mathrm{U}}, \bm{\Sigma}_{\mathrm{V}})
\end{equation}
and which is parameterized by a row (spatial) covariance matrix $\bm{\Sigma}_{\mathrm{U}} \in \mathbb{R}^{C \times C} $ and column (temporal) covariance matrix $\bm{\Sigma}_{\mathrm{V}} \in \mathbb{R}^{K \times K} $. The latter we choose as the covariance matrix of a GRW process, which is defined as $\bm{\Sigma}_{\mathrm{V}} = \sigma_0^2 \bm{1} \bm{1}^\top + \sigma_i^2 \bm{J} \bm{J}^\top$ where $\bm{1}$ is the vector of all ones of length $K$ and $\bm{J}$ is the $K \times K$ unit lower-triangular matrix of ones. The spatial covariance matrix $\bm{\Sigma}_{\mathrm{U}}$ is modeled as the sum of the scalp forward projections of per-source covariance matrices $\bm{\Gamma}_s \in \mathbb{R}^{3 \times 3}$ and a background term $\bm{\Sigma}_{\epsilon}  \in \mathbb{R}^{C \times C}$, and is therefore:

\begin{equation} \label{eq:Sigma_U}
\bm{\Sigma}_{\mathrm{U}} = \bm{\Sigma}_{\epsilon} + \sum_{s=1}^{S} \bm{L}_s \bm{\Gamma}_s \bm{L}_s^\top.
\end{equation}

where $\bm{L} \in \mathbb{R}^{C \times 3S}$ is an EEG lead-field matrix with flexible (3-axis) orientations and $\bm{L}_s$ is the triplet of columns corresponding to source $s$.

The spatial component is essentially the same prior employed by the Champagne M/EEG source estimation model \cite{wipf2010robust} while the overall spatio-temporal construction for $\bm{A}$ is an instance of the Dugh framework \cite{hashemi2021dugh} (a spatio-temporal generalization of Champagne with learnable temporal dynamics), here using the GRW process for the temporal covariance.

We then connect the latent contrast matrix $\bm{A}$ to the decoding weight matrix $\bm{W}^{\mathrm{EEG}}$ via the Haufe transform \cite{haufe2014interpretation} (eq.\ 6) $\bm{A}=\bm{\Sigma}_{X}\bm{W}\kappa$, which we rearrange and simplify to yield $\bm{W}^{\mathrm{EEG}}=\bm{\Sigma}^{-1}_{X}\bm{A}$. The resulting weight matrix can then be viewed as a type of spatial filter that is designed to pick up the contrast encoded in $\bm{A}$ while suppressing noise and interference captured in the empirical data covariance matrix $\bm{\Sigma}_{X}$. This model is also closely connected to linear discriminant analysis (LDA), where $\bm{A}$ plays the role of a difference in per-condition means (the contrast) and $\bm{\Sigma}_{X}$ is the covariance matrix. The latter differs from the average class-conditional covariance matrix $\bm{\Sigma}_{\bar{c}}$ used in LDA in that the covariance of $\bm{A}$ contributes to $\bm{\Sigma}_{X}$ but not $\bm{\Sigma}_{\bar{c}}$, but the effect only impacts the output scale and is absorbed into the scale term $\kappa$ in the Haufe transform (as discussed in \cite{haufe2014interpretation}). This parameter can be made learnable in the Bayesian model, but in our testing this made little to no difference in practice since the scale can also be learned as part of $\bm{A}$ itself, and is therefore omitted from the above formulation.

The main novel aspect is the transformation of what is typically taken as a sparse source imaging model into a sparse discriminant model with the help of the data covariance matrix $\bm{\Sigma}_{X}$.

The lead-field matrix $\bm{L}$ in the above formulation was derived from the ICBM 152-participant average T1-weighted scan \cite{mazziotta2001icbm} from which we reconstructed a 4-shell brain model using Freesurfer \cite{fischl2002freesurfer}. This was then co-registered with the 68-region Desikan-Killiany atlas \cite{desikan2006automated} and the 64 channel BioSemi montage using BrainStorm \cite{tadel2011brainstorm} and was used to estimate an initial 5003-vertex flexible-orientation lead-field matrix. The lead-field matrix $\bm{L}$ as used in the model was then preprocessed by normalizing each row to unit norm and kernel-smoothing as done in the Smooth Champagne variant \cite{cai2019smooth} (here applied to vertices rather than voxels), using a kernel width set to $m=2$ times average distance between adjacent vertices \cite{cai2019smooth} (default is 2 voxels). Also as in Smooth Champagne, the $\bm{\Gamma}_s$ hyper-parameters were tied across source vertices in the same atlas region and parameterized to have spherical covariance, that is, $\bm{\Gamma}_s = \gamma_r^2\bm{I}_3$ where $\bm{I}_3$ is the $3 \times 3$ identity matrix and $\gamma_r^2$ is the scalar variance of the region $r$ into which vertex $s$ falls.

As in most Champagne variants, the vector $\bm{\gamma}$ of hyper-parameters was estimated using empirical Bayes, although here estimation was done as part of the stochastic variational inference (SVI) for the overall model, instead of the iterative closed-form updates prevalent in prior literature, ultimately because our multi-modal decoding model is considerably more complex than these source imaging techniques. As discussed in \cite{wipf2009unified}, the sparse Bayesian learning formulation underlying Champagne can be instantiated using a variety of hyper-priors for the $\gamma_r$ parameters, and flat hyper-priors tend to be preferred since they simplify closed-form updates. Since we are under no such constraints in the SVI framework, we use in our model a weakly informative (half-) student-T hyper-prior with $\nu=3$ degrees of freedom and scale $0.1$\footnote{Wipf and Nagarajan \cite{wipf2009unified} argue that the prior choice is relatively inconsequential, and we find this too, in that the prior type (e.g., flat vs.\ StudentT) and scale had almost no effect on decoding performance in prior exploration on D-BIAT data; however, the sparsity and thus interpretability of the weight map was affected, and thus we used non-flat priors in this work.}:
\begin{equation}
\gamma_r \sim \mathrm{HalfStudentT}(\nu, 0.1),\; \gamma_r \in \mathbb{R}_{+}
\end{equation}

We model the per-modality weights $\alpha_m$ and GRW innovation scale $\sigma^m_\mathrm{i}$ as follows:
\begin{align}
\alpha_{m} &\sim \mathrm{Normal}(0.0, 1.0),\; \alpha_{m} \in \mathbb{R}\\
\sigma^m_\mathrm{i} &\sim \mathrm{HalfNormal}(0.1),\; \sigma^m_\mathrm{i} \in \mathbb{R}_{+}
\end{align}
Our prior analysis of the D-BIAT task had suggested an even tighter hyper-prior with a scale of 0.01 for $\sigma^m_\mathrm{i}$, but this seemed unreasonably rigid for a generic method and was thus relaxed to 0.1 as in the above formulation. We however include a side-by-side analysis of the two parameter choices.

\begin{figure}[htbp]
\centering
\includegraphics[width=\textwidth,trim=4cm 2.5cm 4cm 2.5cm,clip]{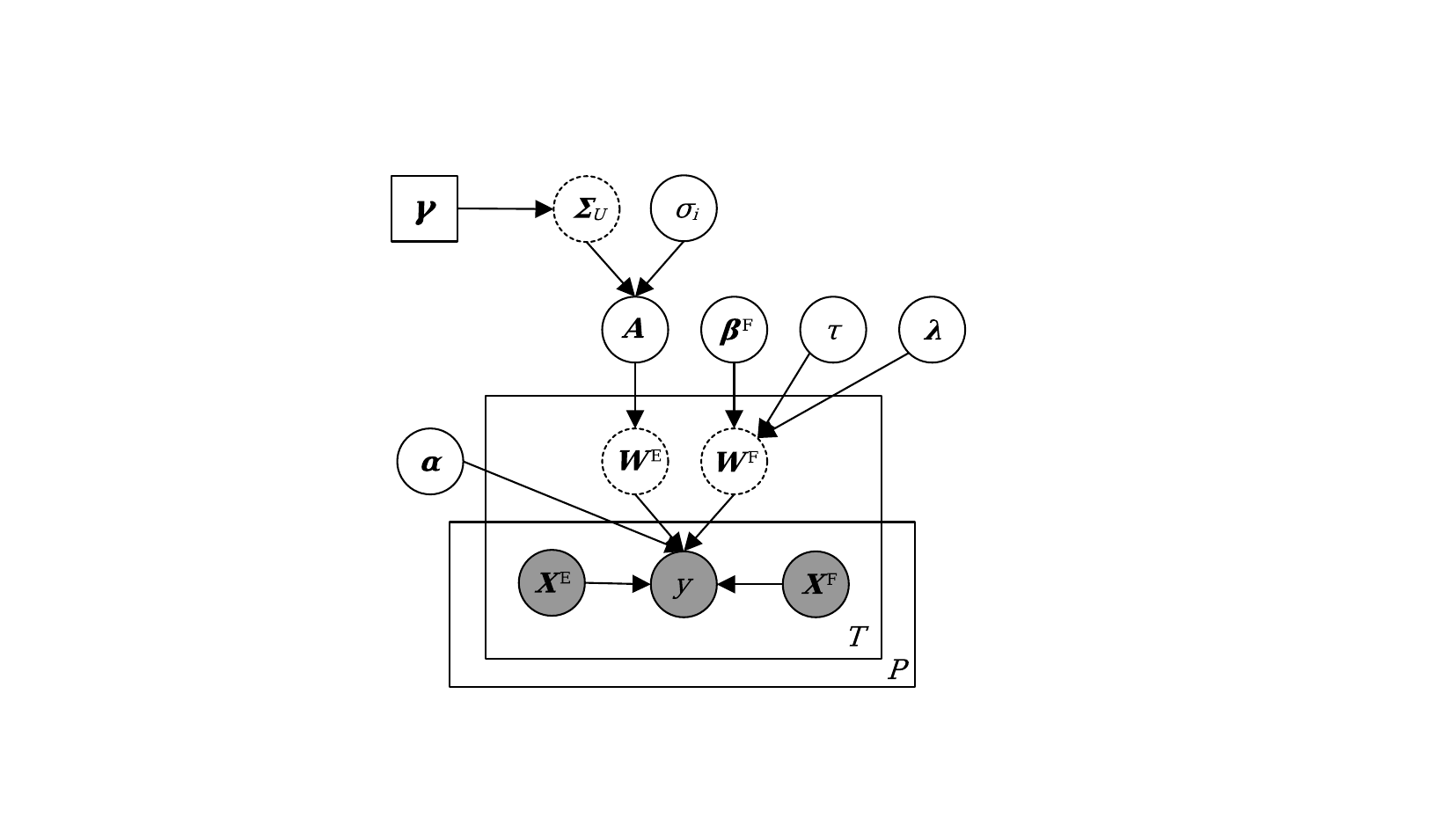}
\caption{
Diagram for the USBL model, shown for a combination of EEG and FAU modalities. Here $T$ is the number of trials in a session and $P$ is the number of participants in the dataset being analyzed. The modality superscripts EEG and FAU have been shortened to E and F, respectively. $\bm{X}^{\mathrm{E}}$ and $\bm{X}^{\mathrm{F}}$ are the preprocessed observations for the current trial and participant (subscripts omitted) and $\bm{W}^{\mathrm{E}}$ and $\bm{W}^{\mathrm{F}}$ are the associated weight matrices. $y$ is the dependent variable for the current participant (same for all trials), $\bm{\alpha}$ is the vector of per-modality weights $\alpha_m$, $\bm{A}$ is the latent EEG contrast matrix, $\bm{\beta}^{\mathrm{F}}$ is the unscaled weight matrix for the FAU modality and $\tau$ and $\bm{\lambda}$ are the global and local (per-channel) horseshoe scales, respectively. $\bm{\Sigma}_{U}$ is the model-derived EEG spatial covariance matrix and $\sigma_i$ is the learned innovation scale for the EEG smoothing (GRW) prior. $\bm{\gamma}$ is the vector of per-region hyper-parameters estimated as part of the model.
}
\label{fig:graphical-model}
\end{figure}

The EEG data covariance matrix was estimated as $\bm{\Sigma}_{X}=(1-\lambda)\bm{\Sigma}_{\mathrm{post}} + \lambda \mathrm{diag}(\bm{\Sigma}_{\mathrm{post}})$ where $\bm{\Sigma}_{\mathrm{post}}$ is the empirical sample covariance matrix of the pooled post-stimulus EEG activity across the training data and $\lambda=0.01$ is a shrinkage regularization parameter, mainly to prevent degeneracy. The noise covariance matrix $\bm{\Sigma}_{\epsilon}$ was in turn estimated from the empirical pre-stimulus covariance matrix $\bm{\Sigma}_{\mathrm{pre}}$ using VBFA based on \cite{nagarajan2007vbfa} (here using 5 interference factors) according to standard Champagne M/EEG practice. Note VBFA is implicitly regularized and therefore $\bm{\Sigma}_{\epsilon}$ does not require a $\lambda$ term. An overview of the complete model is shown in figure~\ref{fig:graphical-model}.

Since the Dugh-type EEG prior is relatively complex and one may wonder if a simpler alternative would be competitive, we also evaluated an alternative formulation that relies on a somewhat more conventional low-rank modeling assumption, where it is assumed that the weight matrix $\bm{W}^{\mathrm{EEG}}$ is low-rank (using a horseshoe prior applied to its singular values), while its right singular vectors follow the same GRW prior as in the other formulations (imposing temporal smoothness); this model performed less well in our evaluation and is only noted here for completeness.

We refer to the overall model throughout the remainder of the article as unified sparse Bayesian learning for IATs (USBL-IAT or short USBL) in that it represents a multi-modal instantiation of SBL that draws on a flexible modality-specific repertoire of sparse priors.

Inference was performed in NeuroPype using the \texttt{numpyro} \cite{phan2019numpyro} probabilistic programming language using its stochastic variational inference \cite{hoffman2013svi} approach in conjunction with the graph-based variational objective \cite{schulman2015graph}, and using a Laplace posterior approximation. For optimization we used the Optax \cite{deepmind2020jax} implementation of Adam \cite{kingma2014adam} with gradient norm clipping at 1.0 and using a learning rate of 0.01 that was exponentially decayed over 5000 steps to a final value of 0.0025. This converges in about 3-5 minutes on an RTX3080-class NVIDIA GPU depending on the combination of modalities included in the model. We did not encounter evidence of local-minima issues with the optimization problem, in that runs from different pseudo-random initializations tended to converge to similar or identical final performance. A full cross-validated evaluation run with/without confidence calibration takes ca. 2.5h/12h on a single GPU per configuration, although note that routine re-fitting of the model on a growing data corpus may not require a full cross-validation and re-evaluation, and a multi-GPU system reduces this by the number of GPUs.

\subsection{Confidence Calibration} \label{sec:confcalib}

Confidence calibration infers the scale factor $\omega$ for the model's logits $z$. The primary reason why this parameter cannot be readily estimated in one step along with the remainder of the model is that our primary observations are at the single-trial level, where the participant-level $z$ is merely a derived variable, resulting in model confidence being calibrated to the evidence available in a single trial, while total evidence across a session worth of trials will typically be larger. Furthermore, if we included an additional participant-level observation and likelihood term in the model to calibrate $\omega$, this would risk considerable overfitting of the overall model at modest numbers of training participants (also empirically observed in prior analysis on B-DIAT data), since the per-participant observation would also influence the high-dimensional weight matrices, among others.

Instead, we fit the model using a two-stage procedure: first the inner model is fit (as described in section \ref{sec:learning}, fixing $\omega=1$), and then $\omega$ is inferred. For the latter we use a strategy similar to Platt scaling \cite{platt1999probabilistic} where we form a posterior predictive distribution $p(z_p)$ over the respective session-level prediction $z_p$ that we evaluate separately for one or more \textit{held-out} participants $p$ as part of an internal cross-validation. As motivated in \cite{platt1999probabilistic}, such a procedure will absorb the generalization error of the method into its confidence. Given the distribution over $z_p$, we model
\begin{align}
\omega &\sim \mathrm{HalfCauchy}(10.0),\; \omega \in \mathbb{R}_{+}\\
y_p &\sim \mathrm{Bernoulli}(\mathrm{logistic}(\omega z_p))
\end{align}

where we pulled $\omega$ out of the formula for $z_p$. This uses a relatively loose prior for $\omega$ that allows the confidence to be increased considerably if supported by the data. Here, $y_p$ is the label of each respective held-out test-set participant. If a full posterior for $\omega$ is inferred, the resulting composite model is then also fully Bayesian. We fit this portion of the model also using \texttt{numpyro}, using Markov-Chain Monte Carlo (MCMC) \cite{gelfand1990sampling} and specifically the No U-Turn Sampler (NUTS) \cite{hoffman2014nuts} using 500 warmup samples and 1000 retained samples.

To avoid potential confusion, we note that, when the resulting composite model-fitting itself is being evaluated in an overall cross-validation (as will be discussed in the following), the latter constitutes an \textit{outer} cross-validation while the aforementioned Platt scaling-``internal'' cross-validation plays the role of a \textit{nested} cross-validation that is rigorously confined to each respective (outer) training set (e.g., \cite{lemm2011introduction}); as this step can multiply the overall compute expense several-fold; for the nested level we use a 5-fold stratified cross-validation (holding out participants).

\subsection{Evaluation} \label{sec:eval}

To assess the performance of our model when generalizing to new participants, we employed a 10x repeated 5-fold randomized and stratified (outer) cross-validation, holding out whole participants\footnote{This amounts to 7--10 participants per fold depending on task (E-IAT or PSY-IAT) and fold-to-fold variation due to rounding.}, in line with current best practices (e.g., \cite{varoquaux2017assessing}, who recommend a similar configuration). We then quantified, on each test set, the AUC as our primary metric along with sensitivity and specificity on the pooled test-set predictions. Notably, when using a decision threshold at $P=0.5$ (or equivalently, $z=0$), the classification decision is invariant under confidence rescaling via $\omega$ and thus metrics such as accuracy, sensitivity, or specificity remain unaffected. The AUC, being sensitive only to ranks, also remains unaffected. For these reasons and for computational expediency, this step was not applied for the main result tables (tables~\ref{tab:R1},~\ref{tab:R2}) and we assumed $\omega = 1$, but full confidence rescaling was applied to generate results in table~\ref{tab:R3}. We tested the AUC of the proposed method for significant difference from a chance level of 0.5 using a corrected resampled t-test, which accounts for the repeated nature of the cross-validation \cite{nadeau1999ttest}. These p-values were further corrected using the Benjamini-Hochberg false discovery rate procedure \cite{benjamini1995controlling} across all 16 instantiations of the proposed method across modalities, tasks, and an additional application to MDD-only participants, but not covering modified variants included to support side remarks in the paper or to illustrate sensitivity to hyper-parameters. In addition to fold-level standard deviations, we report corrected 95\% confidence intervals on the mean performance metrics (tables~\ref{tab:R1},~\ref{tab:R2}), computed using the Nadeau \& Bengio variance correction $\hat{\sigma}^2_{\mathrm{corr}} = (1/(kr) + n_2/n_1) \hat{\sigma}^2$ where $k=5$, $r=10$, and $n_1$, $n_2$ are the training and test set sizes per fold, respectively. This correction accounts for the non-independence of overlapping training sets across folds\footnote{For our specific design parameters, the corrected CI half-width yields CIs that are numerically close to mean SD. This is a coincidence of these particular parameters and does not hold in general.}.

\subsection{Reference Methods} \label{sec:refmethods}

Due to the few applicable reference methods to compare with, we chose comparison baselines as follows. For EEG, we employed a strong linear method (sLDA on linear features, as described in \cite{blankertz2011single}) that is known to work well on tasks that exhibit salient event-related potential (ERP) effects (section Neural Contrast Maps confirms that this is the case here). Additionally we compare with a gold-standard deep learning method (EEGNet \cite{lawhern2018eegnet}) that is understood to be suitable for both ERPs and oscillatory processes. We reasoned that eye-tracking and AU time courses exhibit qualitatively similar time courses as EEG event-related potentials (but playing out over longer time scales), and therefore we use these same methods also for pairings of EEG and non-EEG modalities (suitably extended and using accordingly longer time windows, see Appendix~\ref{app:refmethods} for details), and we use these methods also for the case where EEG is not included.

To shed more light on the effectiveness of the sparse Bayesian priors and the Bayesian procedure in particular (as opposed to regularization and empirical risk minimization), while using the same underlying features, we also compare with a robust (scikit-learn) implementation of $l_2$-regularized logistic regression (shown as L2LR in the results tables). We test this approach on all modality combinations, EEG or otherwise. All methods share the same preprocessing chain as the proposed method.

Notably, all implemented reference methods follow the congruency detection strategy of the Bayesian model (realized by the mirror constraint in the latter and by a label ``recoding'' construction described in Appendix~\ref{app:refmethods}). One may wonder whether \textit{direct} decoding of the participant's label regardless of block type would be possible. In prior work we have generally found, both in the Bayesian model and in the reference methods, that this does not appear to be the case and performance across all methods in this scenario was considerably weaker, often at or near chance level. We consider a full exploration of this route out of scope for this article but include a setup in the comparison matrix to demonstrate this, where we use the single best-performing linear model (PSY-IAT, sLDA, EEG-only) and we adjust the model to directly decode the participant label (thus skipping the construction discussed in Appendix~\ref{app:refmethods}); this approach is identified as ``sLDA (direct)''.

\section{Results} \label{sec:results}

In the following section, we review results across a bank of comparisons, separately for each of the two tasks. We generally show results for all combinations of modalities except where noted (e.g., to save table space where a modality was universally uninformative). For each table cell we reproduce the mean and standard deviation of the respective performance metric across the 10$\times$5 cross-validation folds, corresponding to 50 train/test splits.

\subsection{Decoding Performance} \label{sec:performance}

\subsubsection{E-IAT}

We tested the proposed model across a range of modality subsets to ascertain the individual utility of each modality alone and in combination with others. Results are reproduced in table~\ref{tab:R1} below. Additionally, we compared the method with a set of reference methods from prior literature (bottom of table).

\begin{table}[ht]
\centering
\caption{Performance comparison between the proposed method and a battery of reference methods across different modality combinations, on E-IAT data. Alternative hyper-prior settings are reported in table~\ref{tab:extra}. Note results on FAU and Dyn features in the E-IAT were not significantly different from chance level for the baseline methods (similarly to the proposed method), and are not reproduced in the table for conciseness. Values shown as mean $\pm$ SD across 50 CV folds; bracketed intervals are Nadeau \& Bengio corrected 95\% CIs on the mean. CIs were not truncated at [0, 1] to show nominal coverage rate. For the proposed method, BH-corrected $p$-values ($p_\mathrm{BH}$) are reported below each AUC, corrected across all 16 instantiations of the (unmodified) proposed method (entries in tables~\ref{tab:R1},~\ref{tab:R2}, and~\ref{tab:R4}). Results marginally significant at $q=0.10$ are shown in bold and marked with \dag.}
\label{tab:R1}
\begin{tabular}{ll>{\scriptsize}c>{\scriptsize}c>{\scriptsize}c}
\hline
Method & Modalities & AUC & Sensitivity & Specificity \\
\hline
Proposed & EEG  & \makecell{$0.65 \pm 0.21$ [0.43, 0.86]\\[-2pt]\scriptsize $p_\mathrm{BH}\!=\!0.42$} & $0.64 \pm 0.21$ [0.41, 0.86] & $0.65 \pm 0.25$ [0.39, 0.91] \\
 & \makecell[l]{\textbf{EEG, Gaze}}  & \makecell{$\mathbf{0.73 \pm 0.17}$\textsuperscript{\dag} [0.56, 0.91]\\[-2pt]\scriptsize $p_\mathrm{BH}\!=\!0.053$} & $0.48 \pm 0.24$ [0.23, 0.73] & $0.82 \pm 0.21$ [0.61, 1.04] \\
 & EEG, FAU  & \makecell{$0.48 \pm 0.22$ [0.24, 0.71]\\[-2pt]\scriptsize $p_\mathrm{BH}\!=\!0.88$} & $0.49 \pm 0.23$ [0.24, 0.73] & $0.49 \pm 0.26$ [0.21, 0.77] \\
 & EEG, Dyn  & \makecell{$0.48 \pm 0.23$ [0.24, 0.72]\\[-2pt]\scriptsize $p_\mathrm{BH}\!=\!0.88$} & $0.42 \pm 0.22$ [0.18, 0.65] & $0.52 \pm 0.27$ [0.23, 0.80] \\
 & Gaze  & \makecell{$0.40 \pm 0.21$ [0.17, 0.62]\\[-2pt]\scriptsize $p_\mathrm{BH}\!=\!0.61$} & $0.56 \pm 0.19$ [0.35, 0.76] & $0.33 \pm 0.25$ [0.07, 0.58] \\
 & FAU, Dyn  & \makecell{$0.43 \pm 0.24$ [0.18, 0.69]\\[-2pt]\scriptsize $p_\mathrm{BH}\!=\!0.69$} & $0.40 \pm 0.23$ [0.15, 0.64] & $0.50 \pm 0.27$ [0.21, 0.78] \\
 & FAU, Dyn, Gaze  & \makecell{$0.42 \pm 0.25$ [0.16, 0.69]\\[-2pt]\scriptsize $p_\mathrm{BH}\!=\!0.69$} & $0.31 \pm 0.27$ [0.03, 0.59] & $0.61 \pm 0.25$ [0.35, 0.88] \\
\midrule
D-score & RT & $0.50 \pm 0.23$ [0.26, 0.74] & $0.24 \pm 0.21$ [0.02, 0.46] & $0.91 \pm 0.13$ [0.77, 1.04] \\
L2LR & EEG  & $0.41 \pm 0.21$ [0.18, 0.63] & $0.47 \pm 0.25$ [0.21, 0.74] & $0.36 \pm 0.24$ [0.11, 0.61] \\
sLDA & EEG  & $0.52 \pm 0.23$ [0.27, 0.76] & $0.40 \pm 0.24$ [0.14, 0.65] & $0.55 \pm 0.30$ [0.23, 0.86] \\
EEGNet & EEG  & $0.49 \pm 0.25$ [0.23, 0.75] & $0.55 \pm 0.30$ [0.23, 0.87] & $0.42 \pm 0.35$ [0.05, 0.80] \\
L2LR & EEG, Gaze  & $0.53 \pm 0.24$ [0.28, 0.79] & $0.55 \pm 0.27$ [0.27, 0.83] & $0.48 \pm 0.27$ [0.20, 0.77] \\
sLDA (S) & EEG, Gaze  & $0.50 \pm 0.26$ [0.23, 0.78] & $0.34 \pm 0.25$ [0.08, 0.61] & $0.68 \pm 0.31$ [0.35, 1.01] \\
sLDA (L) & EEG, Gaze  & $0.55 \pm 0.26$ [0.27, 0.82] & $0.37 \pm 0.26$ [0.10, 0.65] & $0.68 \pm 0.31$ [0.35, 1.01] \\
EEGNet & EEG, Gaze  & $0.58 \pm 0.24$ [0.32, 0.84] & $0.44 \pm 0.26$ [0.17, 0.72] & $0.66 \pm 0.32$ [0.33, 0.99] \\
\hline
\end{tabular}
\vspace{2pt}

\noindent{\scriptsize sLDA~(S)/(L): shrinkage LDA with the short (as in EEG) / long (extended) time-window feature set; see Appendix~\ref{app:refmethods}.}
\end{table}

\subsubsection{PSY-IAT}

We performed a similar analysis for the PSY-IAT. Results can be seen in table~\ref{tab:R2}.

\begin{table}[ht]
\centering
\caption{Performance comparison between the proposed method and a battery of reference methods across different modality combinations, on PSY-IAT data. Values shown as mean $\pm$ SD across 50 CV folds; bracketed intervals are Nadeau \& Bengio corrected 95\% CIs on the mean. For the proposed method, BH-corrected $p$-values ($p_\mathrm{BH}$) are reported below each AUC, corrected across the 16 instantiations of the proposed method (entries in tables~\ref{tab:R1},~\ref{tab:R2}, and~\ref{tab:R4}). Results marginally significant at $q=0.10$ are shown in bold and marked with \dag.}
\label{tab:R2}
\footnotesize
\setlength{\tabcolsep}{4pt}
\begin{tabular}{ll>{\scriptsize}c>{\scriptsize}c>{\scriptsize}c}
\hline
Method & Modalities & AUC & Sensitivity & Specificity \\
\hline
Proposed & EEG  & \makecell{$0.62 \pm 0.26$ [0.35, 0.90]\\[-2pt]\scriptsize $p_\mathrm{BH}\!=\!0.61$} & $0.75 \pm 0.22$ [0.51, 0.98] & $0.56 \pm 0.26$ [0.28, 0.83] \\
 & EEG, Gaze  & \makecell{$0.42 \pm 0.25$ [0.16, 0.69]\\[-2pt]\scriptsize $p_\mathrm{BH}\!=\!0.69$} & $0.48 \pm 0.30$ [0.16, 0.80] & $0.47 \pm 0.31$ [0.15, 0.80] \\
 & EEG, FAU, Gaze  & \makecell{$0.67 \pm 0.18$ [0.48, 0.87]\\[-2pt]\scriptsize $p_\mathrm{BH}\!=\!0.22$} & $0.65 \pm 0.20$ [0.43, 0.86] & $0.66 \pm 0.28$ [0.37, 0.96] \\
 & EEG, FAU  & \makecell{$0.69 \pm 0.20$ [0.48, 0.90]\\[-2pt]\scriptsize $p_\mathrm{BH}\!=\!0.22$} & $0.69 \pm 0.22$ [0.46, 0.92] & $0.60 \pm 0.27$ [0.31, 0.89] \\
 & EEG, Dyn  & \makecell{$0.60 \pm 0.21$ [0.38, 0.83]\\[-2pt]\scriptsize $p_\mathrm{BH}\!=\!0.61$} & $0.53 \pm 0.24$ [0.28, 0.78] & $0.58 \pm 0.29$ [0.28, 0.89] \\
 & Gaze  & \makecell{$0.58 \pm 0.23$ [0.35, 0.82]\\[-2pt]\scriptsize $p_\mathrm{BH}\!=\!0.69$} & $0.74 \pm 0.27$ [0.46, 1.02] & $0.39 \pm 0.22$ [0.16, 0.62] \\
 & \makecell[l]{FAU, Dyn} & \makecell{$0.73 \pm 0.19$ [0.53, 0.93]\\[-2pt]\scriptsize $p_\mathrm{BH}\!=\!0.11$} & $0.73 \pm 0.25$ [0.47, 0.99] & $0.65 \pm 0.27$ [0.36, 0.93] \\
 & \makecell[l]{\textbf{FAU, Dyn, Gaze}} & \makecell{$\mathbf{0.76 \pm 0.18}$\textsuperscript{\dag} [0.57, 0.94]\\[-2pt]\scriptsize $p_\mathrm{BH}\!=\!0.053$} & $0.69 \pm 0.22$ [0.46, 0.93] & $0.70 \pm 0.24$ [0.45, 0.95] \\
\midrule
D-score & RT & $0.53 \pm 0.23$ [0.29, 0.77] & $0.41 \pm 0.26$ [0.14, 0.69] & $0.68 \pm 0.26$ [0.40, 0.95] \\
L2LR & EEG  & $0.60 \pm 0.20$ [0.39, 0.80] & $0.72 \pm 0.19$ [0.52, 0.92] & $0.47 \pm 0.25$ [0.21, 0.73] \\
EEGNet & EEG  & $0.64 \pm 0.17$ [0.46, 0.82] & $0.64 \pm 0.20$ [0.43, 0.85] & $0.55 \pm 0.24$ [0.29, 0.81] \\
sLDA & EEG  & $0.76 \pm 0.21$ [0.54, 0.98] & $0.81 \pm 0.17$ [0.63, 0.99] & $0.68 \pm 0.25$ [0.42, 0.95] \\
sLDA (direct) & EEG & $0.44 \pm 0.34$ [0.11, 0.77] & $0.76 \pm 0.28$ [0.49, 1.03] & $0.17 \pm 0.37$ [-0.19, 0.53] \\
L2LR & FAU  & $0.65 \pm 0.27$ [0.36, 0.93] & $0.48 \pm 0.31$ [0.15, 0.81] & $0.68 \pm 0.30$ [0.36, 1.00] \\
sLDA (S) & FAU  & $0.56 \pm 0.23$ [0.31, 0.80] & $0.39 \pm 0.26$ [0.12, 0.66] & $0.71 \pm 0.28$ [0.41, 1.00] \\
sLDA (L) & FAU  & $0.56 \pm 0.24$ [0.31, 0.82] & $0.43 \pm 0.29$ [0.13, 0.73] & $0.70 \pm 0.28$ [0.41, 1.00] \\
EEGNet & FAU  & $0.56 \pm 0.24$ [0.31, 0.82] & $0.47 \pm 0.27$ [0.18, 0.75] & $0.61 \pm 0.30$ [0.30, 0.93] \\
L2LR & Dyn  & $0.57 \pm 0.29$ [0.27, 0.89] & $0.54 \pm 0.30$ [0.22, 0.86] & $0.52 \pm 0.34$ [0.16, 0.87] \\
sLDA (S) & Dyn  & $0.56 \pm 0.26$ [0.28, 0.83] & $0.49 \pm 0.30$ [0.18, 0.81] & $0.49 \pm 0.31$ [0.16, 0.82] \\
sLDA (L) & Dyn  & $0.55 \pm 0.26$ [0.28, 0.82] & $0.46 \pm 0.31$ [0.13, 0.78] & $0.56 \pm 0.28$ [0.27, 0.87] \\
EEGNet & Dyn  & $0.54 \pm 0.28$ [0.25, 0.83] & $0.62 \pm 0.28$ [0.32, 0.92] & $0.51 \pm 0.34$ [0.15, 0.88] \\
L2LR & EEG, FAU  & $0.62 \pm 0.25$ [0.36, 0.88] & $0.51 \pm 0.29$ [0.20, 0.81] & $0.67 \pm 0.28$ [0.38, 0.96] \\
sLDA (S) & EEG, FAU  & $0.61 \pm 0.21$ [0.39, 0.83] & $0.50 \pm 0.27$ [0.21, 0.78] & $0.69 \pm 0.29$ [0.38, 1.00] \\
sLDA (L) & EEG, FAU  & $0.63 \pm 0.22$ [0.40, 0.86] & $0.50 \pm 0.26$ [0.22, 0.78] & $0.70 \pm 0.30$ [0.39, 1.02] \\
EEGNet & EEG, FAU  & $0.58 \pm 0.25$ [0.31, 0.84] & $0.44 \pm 0.34$ [0.08, 0.80] & $0.63 \pm 0.32$ [0.28, 0.97] \\
L2LR & FAU, Dyn  & $0.62 \pm 0.22$ [0.38, 0.85] & $0.61 \pm 0.27$ [0.32, 0.90] & $0.55 \pm 0.29$ [0.25, 0.85] \\
sLDA (S) & FAU, Dyn  & $0.54 \pm 0.24$ [0.28, 0.79] & $0.55 \pm 0.30$ [0.24, 0.87] & $0.57 \pm 0.32$ [0.23, 0.90] \\
sLDA (L) & FAU, Dyn  & $0.55 \pm 0.23$ [0.31, 0.79] & $0.55 \pm 0.31$ [0.22, 0.88] & $0.51 \pm 0.32$ [0.17, 0.84] \\
EEGNet & FAU, Dyn  & $0.57 \pm 0.26$ [0.30, 0.84] & $0.54 \pm 0.31$ [0.21, 0.86] & $0.59 \pm 0.34$ [0.24, 0.95] \\
L2LR & FAU, Dyn, Gaze  & $0.57 \pm 0.21$ [0.34, 0.79] & $0.53 \pm 0.30$ [0.21, 0.85] & $0.55 \pm 0.25$ [0.29, 0.81] \\
sLDA (S) & FAU, Dyn, Gaze  & $0.50 \pm 0.25$ [0.24, 0.76] & $0.52 \pm 0.31$ [0.19, 0.84] & $0.59 \pm 0.33$ [0.24, 0.93] \\
sLDA (L) & FAU, Dyn, Gaze  & $0.51 \pm 0.22$ [0.27, 0.74] & $0.52 \pm 0.32$ [0.19, 0.86] & $0.51 \pm 0.31$ [0.18, 0.84] \\
EEGNet & FAU, Dyn, Gaze  & $0.54 \pm 0.24$ [0.28, 0.80] & $0.55 \pm 0.32$ [0.21, 0.89] & $0.54 \pm 0.32$ [0.20, 0.88] \\
\hline
\end{tabular}
\vspace{2pt}

\noindent{\scriptsize sLDA~(S)/(L): shrinkage LDA with the short / long time-window feature set. ``sLDA~(direct)'': variant decoding participant labels directly without the congruency-detection setup. See Appendix~\ref{app:refmethods}.}
\end{table}

The proposed method achieved parity with the respective best reference method on both tasks; notably, no reference method had consistently good performance on both tasks, while the proposed model did. To formally assess whether the proposed method significantly outperforms the best reference method, we conducted a Nadeau \& Bengio corrected paired $t$-test on the fold-level AUC observations, comparing the best-performing proposed-method configuration (HB2, EEG+Gaze, E-IAT; AUC $= 0.73$) against the strongest reference method on the same modality combination (EEGNet, EEG+Gaze; AUC $= 0.58$). The mean fold-level difference was $0.15$ in favor of the proposed method, but this did not reach statistical significance after the variance correction ($t(49) = 0.77$, $p = 0.44$). This reflects the limited power of the corrected test to detect method differences at the present sample size---the Nadeau \& Bengio correction inflates the standard error by a factor of $\sqrt{(1/k + n_2/n_1) \cdot k} \approx 3.7$ relative to the naive paired $t$-test, making direct head-to-head comparisons between methods particularly conservative.

For the respective best-performing setups on the E-IAT and PSY-IAT tasks, we also reproduce Brier scores \cite{brier1950verification} (essentially the mean-squared error between probabilities and labels) and cross-entropy for the basic and the probability-calibrated variant. Both scores capture how well-calibrated the confidence of the predictions is, in addition to their accuracy, but note that these metrics generally change relatively little between best and worst possible confidence for a given accuracy (see also section \ref{sec:discussion} for additional discussion). Results are reproduced in table~\ref{tab:R3} below.

\begin{table}[ht]
\centering
\caption{Comparison of Brier scores (MSE) and cross-entropy between base and calibrated models, for all modality combinations with highest AUC results. Values shown as mean [95\% CI]. Lower scores indicate better calibration.}
\label{tab:R3}
\begin{tabular}{llccc}
\hline
Task & Modalities & Model Variant & MSE & Cross-Entropy \\
\hline
E-IAT & EEG, Gaze & Base & $0.249$ [$0.247$, $0.250$] & $0.690$ [$0.688$, $0.693$] \\
 &  & Calibrated & $0.228$ [$0.190$, $0.267$] & $0.650$ [$0.562$, $0.739$] \\
\hline
PSY-IAT & FAU, Dyn & Base & $0.242$ [$0.236$, $0.249$] & $0.678$ [$0.665$, $0.691$] \\
 &  & Calibrated & $0.233$ [$0.180$, $0.285$] & $0.685$ [$0.521$, $0.850$] \\
 & FAU, Dyn, Gaze & Base & $0.242$ [$0.236$, $0.248$] & $0.677$ [$0.666$, $0.688$] \\
 &  & Calibrated & $0.227$ [$0.180$, $0.275$] & $0.667$ [$0.527$, $0.807$] \\
\hline
\end{tabular}
\end{table}

In table~\ref{tab:extra} we further explore the sensitivity of the proposed method to a number of parameter and modeling choices. Note these analyses were not used to adapt parameters of the method to the dataset, and are presented here for informational purposes.

\begin{table}[ht]
\centering
\caption{Supplementary parameter variations of the proposed method, comparing an alternative smoothness hyper-prior ($\sigma$=0.01; see discussion) and a simpler Bayesian low-rank (LR) model variant for EEG. Values shown as mean $\pm$ SD across 50 CV folds; bracketed intervals are Nadeau \& Bengio corrected 95\% CIs on the mean.}
\label{tab:extra}
\begin{tabular}{ll>{\scriptsize}c>{\scriptsize}c>{\scriptsize}c}
\hline
Task / Modalities & Variant & AUC & Sensitivity & Specificity \\
\hline
\makecell[l]{E-IAT\\EEG, Gaze} & \makecell{$\sigma$=0.1\\(default)} & $0.73 \pm 0.17$ [0.56, 0.91] & $0.48 \pm 0.24$ [0.23, 0.73] & $0.82 \pm 0.21$ [0.61, 1.04] \\
\makecell[l]{E-IAT\\EEG, Gaze} & $\sigma$=0.01 & $0.77 \pm 0.17$ [0.59, 0.95] & $0.48 \pm 0.23$ [0.23, 0.72] & $0.85 \pm 0.20$ [0.64, 1.05] \\
\hline
\makecell[l]{PSY-IAT\\EEG} & \makecell{Dugh\\(default)} & $0.62 \pm 0.26$ [0.35, 0.90] & $0.75 \pm 0.22$ [0.51, 0.98] & $0.56 \pm 0.26$ [0.28, 0.83] \\
\makecell[l]{PSY-IAT\\EEG} & LR & $0.56 \pm 0.20$ [0.35, 0.78] & $0.53 \pm 0.22$ [0.30, 0.76] & $0.48 \pm 0.29$ [0.18, 0.79] \\
\makecell[l]{PSY-IAT\\FAU, Dyn} & \makecell{$\sigma$=0.1\\(default)} & $0.73 \pm 0.19$ [0.53, 0.93] & $0.73 \pm 0.25$ [0.47, 0.99] & $0.65 \pm 0.27$ [0.36, 0.93] \\
\makecell[l]{PSY-IAT\\FAU, Dyn} & $\sigma$=0.01 & $0.71 \pm 0.20$ [0.49, 0.92] & $0.71 \pm 0.25$ [0.44, 0.97] & $0.61 \pm 0.29$ [0.30, 0.91] \\
\makecell[l]{PSY-IAT\\FAU, Dyn, Gaze} & \makecell{$\sigma$=0.1\\(default)} & $0.76 \pm 0.18$ [0.57, 0.94] & $0.69 \pm 0.22$ [0.46, 0.93] & $0.70 \pm 0.24$ [0.45, 0.95] \\
\makecell[l]{PSY-IAT\\FAU, Dyn, Gaze} & $\sigma$=0.01 & $0.61 \pm 0.19$ [0.42, 0.81] & $0.57 \pm 0.24$ [0.32, 0.82] & $0.57 \pm 0.27$ [0.28, 0.85] \\
\hline
\end{tabular}
\end{table}

\begin{table}[ht]
\centering
\caption{Performance of the proposed method on the E-IAT when restricting the analysis to MDD participants only ($n=33$), excluding the 6 control subjects. This addresses the concern (see section Ground Truthing) that our model may partly be detecting an MDD/CTL distinction rather than the entrapment contrast. Values shown as mean $\pm$ SD across 50 CV folds; bracketed intervals are Nadeau \& Bengio corrected 95\% CIs on the mean. The default-$\sigma$ configuration is included in the BH-FDR family jointly with tables~\ref{tab:R1}/\ref{tab:R2}; the alternative $\sigma$=0.01 variant is included for informational purposes and was not enrolled in statistical tests. AUC results significant after BH correction ($p_\mathrm{BH}<0.05$) are shown in bold and marked with *.}
\label{tab:R4}
\begin{tabular}{ll>{\scriptsize}c>{\scriptsize}c>{\scriptsize}c}
\hline
Method & Modalities & AUC & Sensitivity & Specificity \\
\hline
Proposed & \makecell[l]{\textbf{EEG, Gaze}} & \makecell{$\mathbf{0.79 \pm 0.17}$* [0.62, 0.97]\\[-2pt]\scriptsize $p_\mathrm{BH}\!=\!0.032$} & $0.47 \pm 0.23$ [0.22, 0.71] & $0.82 \pm 0.21$ [0.60, 1.05] \\
 & \makecell[l]{EEG, Gaze\\($\sigma$=0.01)} & $0.84 \pm 0.15$ [0.68, 1.00] & $0.44 \pm 0.21$ [0.22, 0.67] & $0.91 \pm 0.18$ [0.72, 1.09] \\
\hline
\end{tabular}
\end{table}

As noted we performed an analysis of the E-IAT restricted to MDD-only participants. This analysis (table~\ref{tab:R4}) yields the strongest E-IAT result observed in this study, with an AUC of 0.84 [0.68, 1.00] for EEG+Gaze ($\sigma=0.01$); the default-$\sigma$ configuration (AUC 0.79) is the only result in this study to survive BH-FDR correction at the conventional $q=0.05$ threshold ($p_\mathrm{BH}=0.032$). This result is important for two reasons: first, it rules out the possibility that the model is primarily detecting the MDD/CTL group distinction rather than the entrapment contrast; and second, the MDD-only scenario is arguably more representative of a setting where subjects with diagnosed MDD are participating in an E-IAT along with other downstream diagnostic tools, e.g., in a hypothetical future clinical evaluation protocol.

\subsection{Model Weights} \label{sec:weights}

The following section presents model weights for the best-performing models along with technical notes on their structure.

\begin{figure}[htbp]
\centering
($a$)\\[2pt]
\includegraphics[width=\figwidth]{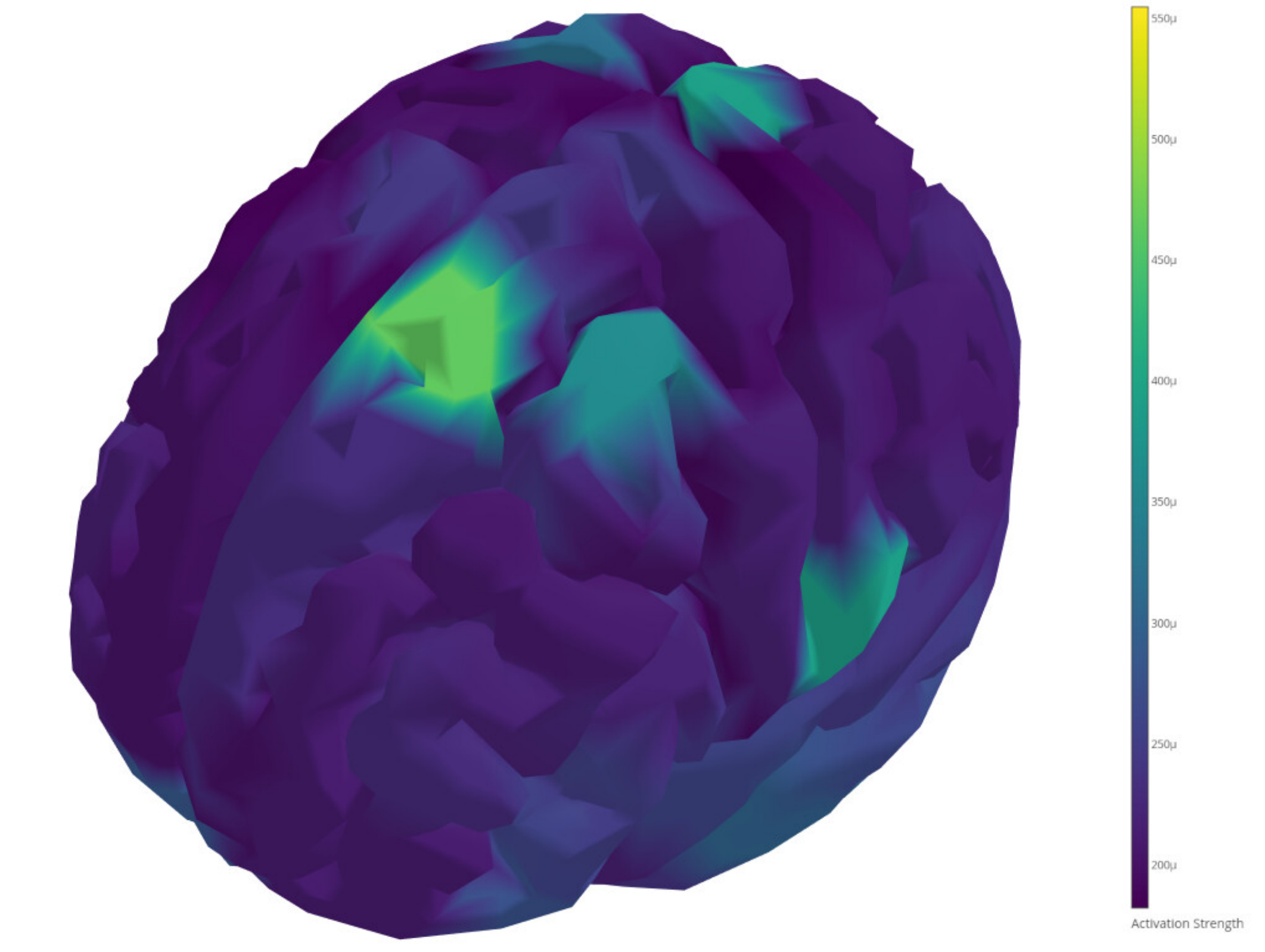}

\vspace{6pt}

($b$)\\[2pt]
\includegraphics[width=\figwidth]{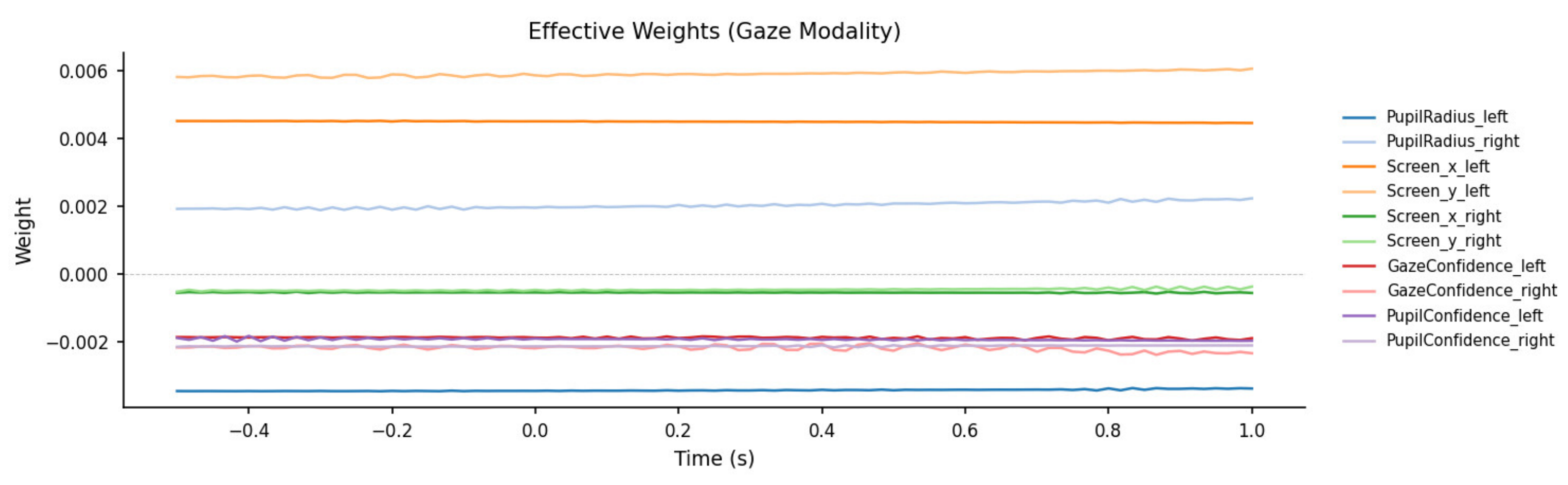}
\caption{EEG and eye-tracking weights for the best-performing E-IAT model; ($a$) spatial EEG weight maps, ($b$) eye tracking weight time course (note 0.0 is not at the center of the vertical axis).}
\label{fig:weights-eiat}
\end{figure}

The EEG component of these models is natively represented in brain space, where the $\bm{\gamma}$ hyper-parameters can be interpreted as inferred variance (activity level) in a contrast map (difference between congruent and incongruent) in each region of the brain. Note that these variance parameters are non-negative and thus do not encode the directionality of the effect, only its magnitude. Figure~\ref{fig:weights-eiat} shows EEG weight hotspots learned by this model (averaged across time in epoch), which are natively represented in cortical space owing to the construction of the model. The weight maps show hotspots in left medial frontal cortex (highest activation), caudal medial frontal cortex (overlapping with Brodman Area 6), and left lateral primary motor cortex. The constellation overlaps with regions associated with conflict detection (Anterior Cingulate Cortex / ACC; also noted in \cite{chee2000dorsolateral}), sensori-motor areas (SMA), and executive control areas mediating between the two (Brodman Area 6). However, although this is well aligned with conflict theory as applied to IATs, we caution against drawing strong anatomical conclusions based on these data since the model is fundamentally a statistical estimate, and the accuracy of our source estimation is also somewhat limited by the lack of individual-specific anatomical ground truth (i.e., individual T1-weighted MRI scans) or template warping.

The associated eye tracking weight shows that the model mainly relies on the time average (over the course of a time-locked segment) of the eye pose without detailed time-locking to the stimulus onset; the most informative parameters were predominantly the left-eye screen coordinates, left pupil radius, and gaze confidence (which can act as a proxy for the absence of blinks). The relatively flat time course is likely due to the model not being able to explain a more detailed time structure given the amount of training data available, causing the GRW innovation scale being shrunk to close to zero; another explanation may be a high degree of variability in eye responses leading to only the (block-design imposed) condition average being informative.

In the best-performing PSY-IAT model (FAU+Dyn), we also observed that the weight time course tended to not change much over the course of a trial, implying that the model primarily learns average FAU states that appear to remain near-constant across the block, rather than temporal reactions to each individual stimulus, likely for the same reason as in the E-IAT gaze dynamics. In contrast, the weight matrix for the Dyn stream (holding temporal derivatives), which is not subject to a smoothness prior, had no discernible structure (Figure~\ref{fig:weights-psyiat} in supplement).

\subsection{Probability Calibration} \label{sec:probcal}

Probability calibration, when used, led to improved Brier scores (MSE) on both tasks (table~\ref{tab:R3}). For the E-IAT, the MSE improved from $0.249$ to $0.228$ and cross-entropy from $0.690$ to $0.650$.  For the PSY-IAT, the base MSE of $0.242$ likewise improved modestly to $0.227$. The range for the Brier score is rather narrow, in that $0.25$ corresponds to the uninformative-prediction Brier score (all predictions at ${\sim}0.5$), and the expected (best-possible) MSE at an AUC/balanced accuracy of ${\sim}0.7$ is approximately 0.21; in that sense the improvements cross more than half of the possible range in each of the tasks. Confidence intervals of the calibrated method are somewhat wider due to the $\omega$ parameter being calibrated on fewer data points than the remainder of the model.

\subsection{Neural Contrast Maps} \label{sec:neural}

In the following we review topographic maps of neural condition differences per group by task along with event-related potentials to shed more light on the neural dynamics underlying these models.

\subsubsection{E-IAT}

\begin{figure}[htbp]
\noindent\includegraphics[width=\figwidth]{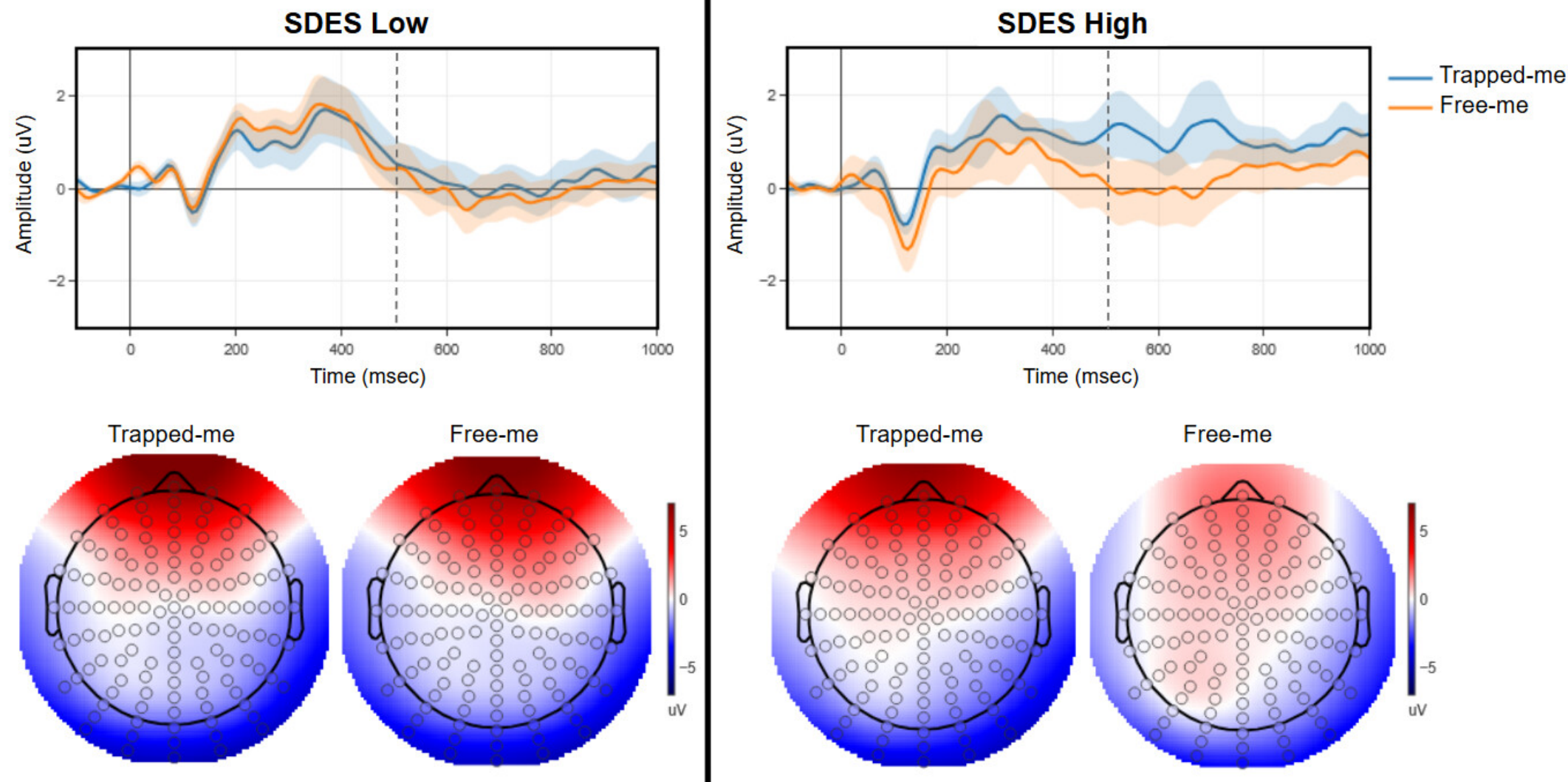}
\caption{Mean ERP at channel F7 (top) by group (left=low entrapment, right=high entrapment) for each condition (``trapped/me'' vs ``free/me'') with the scalp topographies underneath at time $t=500$ msec after stimulus onset.}
\label{fig:neural-eiat}
\end{figure}

The mean ERPs showed the greatest group by condition differences at the left frontal region (channel F7) in the late component time periods $t=400$--$600$ ms after stimulus onset. For the low entrapment group (``SDES Low'' in the figure) there were very few differences between the ``trapped/me'' and ``free/me'' conditions, while the high entrapment group (SDES High) showed increased EEG amplitudes in the frontal areas for the ``trapped/me'' condition (figure~\ref{fig:neural-eiat}).

\subsubsection{PSY-IAT}

\begin{figure}[htbp]
\noindent\includegraphics[width=\figwidth]{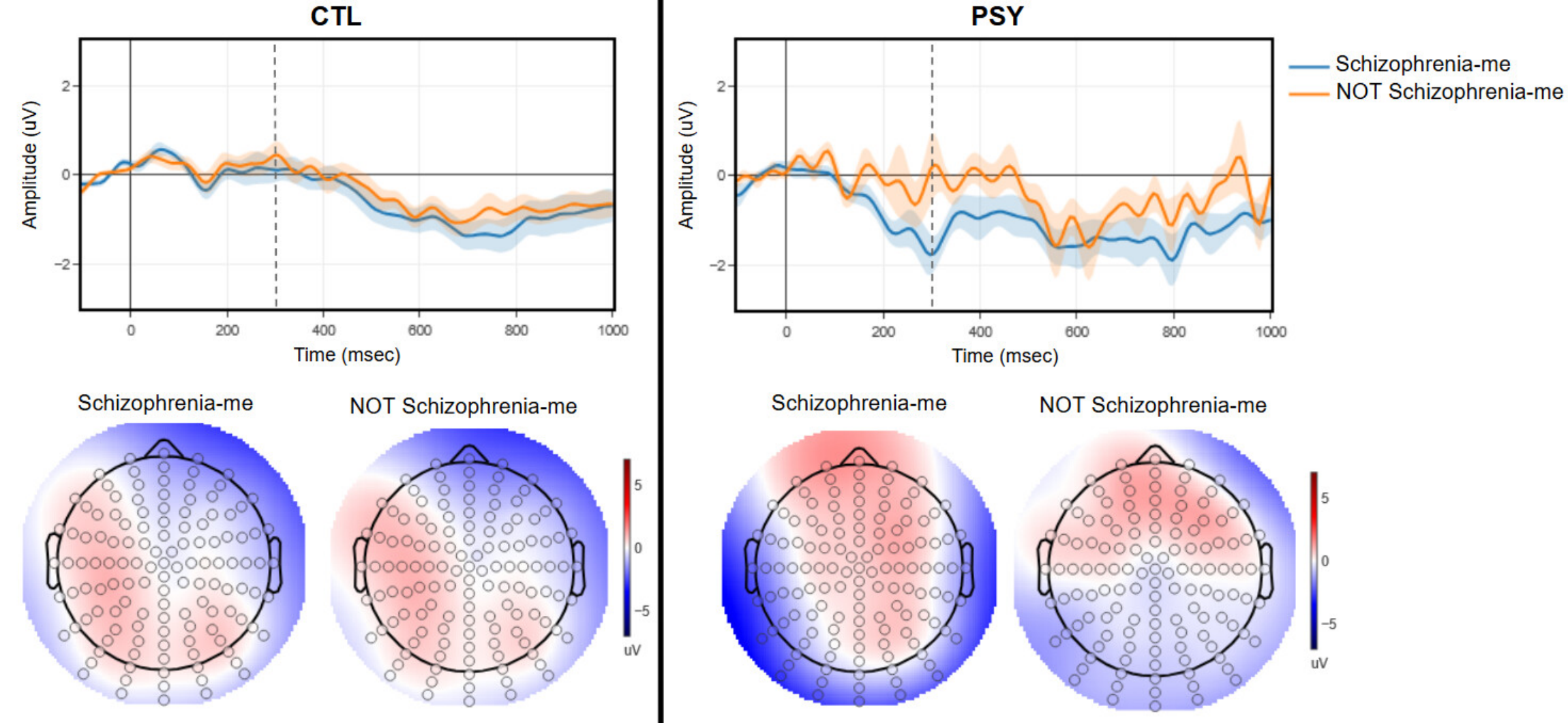}
\caption{Mean ERP at channel T7 (top) by group (left=control, right=psychosis) for each condition (``Schizophrenia/me'' vs ``NOT Schizophrenia/me'') with the scalp topographies underneath at time $t=300$ msec after stimulus onset.}
\label{fig:neural-psyiat}
\end{figure}

For the PSY-IAT, the mean ERP results showed similar patterns, but at different brain regions and at different time periods than for the E-IAT. Here we found for the control group little to no differences between conditions ``Schizophrenia/me'' and ``NOT Schizophrenia/me'', while the PSY group shows increased P300 amplitudes in the left temporal region for the ``Schizophrenia/me'' condition (figure~\ref{fig:neural-psyiat}).

\section{Discussion} \label{sec:discussion}

\textit{Performance.} Our results suggest that the proposed method can predict mental health-related variables at the session level, with point-estimate AUCs in the 0.73--0.76 range for the best-performing configurations, although the corrected confidence intervals remain wide, and full-sample results do not survive false discovery rate correction at $q=0.05$ (see tables~\ref{tab:R1}/\ref{tab:R2}, with $p_{\mathrm{BH}}=0.053$), but are marginally significant at $q=0.10$. Performance varies substantially across modality combinations. For the E-IAT, we reach a respectable performance of 0.73 AUC when predicting binarized entrapment from the combination of EEG and eye-tracking modalities (0.77 with the tighter $\sigma=0.01$  hyper-prior). The traditional D-score method, by contrast, is at chance level on the E-IAT (AUC=0.50, CI [0.26, 0.74]) and exhibits a strongly imbalanced operating point with very low sensitivity (0.24) and high specificity (0.91), while the proposed method is generally better balanced across all tasks and modalities. When restricting the E-IAT analysis to MDD participants only, performance improves to AUC 0.79 (table~\ref{tab:R4}), with a corrected 95\% CI lower bound of 0.62; this is the strongest result in this study and the only configuration to survive BH-FDR correction at the conventional $q=0.05$ threshold ($p_\mathrm{BH}=0.032$). For PSY-IAT, we find that purely behavioral measures (FAU+Dyn+Gaze) alone yield strong performance (0.76 AUC), and this outperforms setups that include EEG, likely due to overfitting challenges given the modest dataset size. Here, the D-score method is also at near-chance (AUC 0.53); the latter is considerably lower than in the original introduction of this task \cite{kirschenbaum2022validation}, and might point at differences in participant cohort or possibly differences due to the brief task variant used here.

To place these results in context: the D-score remains the predominant analytic method for IATs in both research and applied settings, and on both tasks studied here it produced near-chance classification. The point-estimate improvements of 0.23 AUC over D-scores on each task are substantial, even though the multiplicity correction applied across all 16 tested configurations of the proposed method---spanning two tasks and multiple modality combinations---limits the formal statistical significance of individual comparisons. This testing regime was chosen to transparently report the full space of results rather than to optimize for apparent significance, and reflects genuine prior uncertainty about which modality combinations would prove most informative.

As noted in section~\ref{sec:related}, there are few directly comparable ML results on multi-modal IAT data, so we compared against generic off-the-shelf approaches. Among these, for the PSY-IAT, sLDA on EEG-only data matched USBL's best performance on behavioral data (0.76 AUC) but fell short on all other modality combinations and on the E-IAT (by 0.10--0.20 AUC). EEGNet was the best among reference methods on the E-IAT but was numerically exceeded by the Bayesian method (by 0.15 AUC on the best-matched modality combination, though the head-to-head paired comparison did not reach statistical significance). In the PSY-IAT EEGNet reached similar performance to the Bayesian method on EEG-only but fell short on other modality combinations (by as much as 0.22 AUC) . Deep learning methods also proved difficult to make competitive on FAU data despite exploration of EEGNet, HTnet \cite{peterson2021generalized}, and multi-modal extensions thereof. Taken together, no single reference method matched USBL on both tasks, despite reference methods having been modestly tuned to the tasks at hand, whereas USBL has few user-adjustable parameters (mainly the analysis time window and the smoothness prior scale).

\textit{Flexibility and Parameterization.} The fact that our models achieve competitive performance when the number and nature of information-carrying modalities changes speaks to the versatility of the framework. Our method's flexibility with respect to modality is desirable because it supports application to IAT studies with different measurement capabilities and allows for the possibility that different sensor streams may be warranted under various study goals or constraints. The method can, we believe, be adapted to a wider range of IAT task types than tested here, with early concurrent work on the D-BIAT providing some evidence \cite{bronstein2025neurobehavioral}.

Applying USBL to yet other IATs in principle merely requires plugging in the respective dependent variable and potentially choosing a time window of interest (which was not varied between the two tasks explored here). This time window may in theory depend on the nature and complexity of stimuli being used (e.g., word length and familiarity) and is thus a free parameter of the method. The other main choice is the type of prior per modality, although this is a largely mechanical choice: Dugh-type for EEG, low-rank for general spatio-temporally correlated variables, group-sparse for modalities with relatively independent channels (here FAU, eye tracker), pointwise sparse for modalities with many independent features of which only few are believed relevant (e.g., genomic factors, tabular features, etc), with a possible Gaussian prior when none of these cases apply (or for scalar variables such as reaction times).

In an attempt to assess the sensitivity of the method to the temporal smoothness parameter (GRW innovation scale) we provided results with both settings side by side, for the best-performing setup on each task (table~\ref{tab:extra}). This shows that the tighter prior improves performance somewhat for the E-IAT while reducing it for the PSY-IAT, and confirms that the method is somewhat sensitive to this parameter, at least in the regime of relatively small datasets. A second area where this choice has an effect is in the model weights, which are rather flat over time (e.g., figure~\ref{fig:weights-psyiat}); we found that this continued to be the case even with a very lax value of $\sigma=1.0$ (not shown), suggesting that this effect is indeed driven by the data rather than imposed by the choice of hyperprior.

\textit{Parameter Efficiency.} The sparse Bayesian approach employed in our USBL framework was ultimately motivated by a suspicion that the critical missing ingredient in a diagnostic-like system driven by high-dimensional input signals, and in the absence of heavy feature engineering, may lie in parameter efficiency of the model. Machine learning on trial-oriented computer-based cognitive tasks is, generally speaking, a well-trodden regime, and it is often assumed (for example in EEG, as in \cite{blankertz2011single}), that variability across trials is relatively high and variability across participants is modest enough to be surmountable, this permitting cross-participant generalization. In such a setting, the sample size behaves essentially like the total number of trials across all participants, which will often be in the thousands or tens of thousands, and this in turn allows relatively complex models to be fit. However, when inter-individual variability is relatively high (e.g., relative to effect size or within-session trial-to-trial variability), then the effective sample size will be closer to the total number of training participants, which is orders of magnitude lower (e.g., 30--40), and that in turn would require models whose effective degrees of freedom is considerably lower than the number of available input features, which leads us to sparse models.

The consequence was to focus on maximally parameter-efficient models without requiring user-tunable features or hyper-parameters, which suggested models that adapt to the data at hand in their effective capacity by means of (implicit) Bayesian model averaging or selection, and specifically sparse Bayesian models. While sparsity is straightforward for data that is sparse in obvious respects, e.g., relevant features, and still somewhat straightforward when working with variables that admit a natural grouping structure, such as FAUs or eye-tracker channels, EEG data does not admit a simple sparsity structure, since information is spread out across many scalp channels due to brain and skull volume conduction, which prompted the new method design.

\textit{Interpretation.} The present study provides some insight into the likely informativeness of different sensor modalities with respect to the IAT outcome variable, subject to the assumptions of our (generalized) linear USBL model. Broadly speaking, in our study the high-dimensional modalities (EEG, FAU) were considerably more informative than the relatively lower-dimensional eye tracking time series. Eye tracking all by itself yielded an AUC of 0.58 for the PSY-IAT and 0.40 for the E-IAT. It is not obvious why, in the latter task, inclusion of eye-tracking data improved performance when combined with EEG over EEG alone, and this is not easy to disentangle even in a linear decoding model; however, the same effect was also seen in EEGNet. One possibility is that the eye-tracking signal may have acted as a sort of control variate with respect to the EEG signal (e.g., reducing variability due to whether stimuli were looked at or not). Another possibility is that the mere presence of the extra variables forced the sparsity terms in the model to eliminate or rebalance model weight across the EEG features, and that could plausibly suppress weight on nuisance features in a way that may ultimately result in a better-generalizing model.

\textit{Calibrated confidence.} We confirmed a meaningful improvement in model confidence as a result of the Bayesian confidence calibration, as seen in the analyses of Brier scores (table~\ref{tab:R3}, which improved from near-chance half-way towards the best value expected at an AUC or accuracy of approximately 0.7). This would be useful in future clinical or public health settings, where medical practitioners may factor the predicted probability into their own decision making. In this context, we consider it a feature that the methodology is fully Bayesian and in that sense aligns with the Bayesian probabilistic framework increasingly advocated in clinical decision-making. Nevertheless, we acknowledge that the present model, and decoding from complex measures such as EEG or behavioral time series, remains ultimately difficult to interpret, despite the ability to inspect model weights.

\textit{Limitations.} Several limitations should be noted. The E-IAT and its EMA-derived entrapment dependent variable are novel instruments whose psychometric properties---including test-retest reliability and concurrent validity against established entrapment measures---are being investigated in separate ongoing work. As such, the E-IAT results presented here should be understood primarily as a demonstration of the method's generality across a second IAT variant with a different type of dependent variable, rather than as a definitive validation of the entrapment construct itself. If the EMA-derived score were unreliable or poorly operationalized, this would be expected to attenuate (rather than inflate) observed classification performance, and the above-chance results therefore provide indirect evidence that the construct carries meaningful signal. The PSY-IAT, grounded in established clinical diagnosis, provides the more conservative benchmark for evaluating method performance. Additional limitations include the modest sample sizes discussed above and the single-site data collection for each task, which leave generalization to other hardware configurations, populations, or clinical definitions an open problem.

\section{Conclusion} \label{sec:conclusion}

We presented a novel approach for inferring the primary outcome variable from multi-modal time-series data collected in parallel during IAT task performance, showing performance competitive with and more consistent than competing approaches, specifically the prevailing D-score method.

\textit{Innovations.} A foundational innovation in our approach was the differential (congruency-detection) setup, both in the Bayesian model and in how conventional off-the-shelf models were applied to the problem (Appendix~\ref{app:refmethods}). Incorporating EEG data into the framework required adapting a state of the art sparse source estimation method (Champagne and its various generalizations, including the Dugh framework) and recasting it into a spatial filter design approach by means of the Haufe transform. This yielded an elegant and demonstrably competitive solution for EEG that composes naturally with the other types of sparsity terms (such as groupwise sparsity or pointwise sparsity) in other modalities or auxiliary (tabular) variables.

The second prior structure that we made consistent use of is a smoothness assumption over time, which is often justified in biological time-series data; in our model this is generally realized with a Gaussian Random Walk (GRW) prior, and this can be flexibly combined with all aforementioned types of priors, except pointwise sparsity. Besides the sparsity terms, this was the second ``unlock'' that made both high-dimensional EEG and high-dimensional FAU or gaze analysis fully tractable on our datasets, and underpins, along with sparsity, the respectively best results on E-IAT and PSY-IAT presented above.

\textit{Future Work.} The present study suggests several areas of future exploration. One unanswered question is what combination of modalities best enables IAT task decoding, as this was somewhat inconsistent across the tasks compared here. Another major component of future work would be the partial adaptation of the empirical noise covariance matrix (in the Haufe transform) to individual participants. Other elements of the model that could potentially be improved include a multiplicative term per stimulus type (word) that models the salience or informativeness of that word relative to other words; fitting these extra terms however would likely require more data than we had available so this was not explored. Likewise, one might model a time-on-task effect or a habituation effect within-block; these components were likewise deferred due to the extra model degrees of freedom that they introduce, which in turn can only be estimated with larger cohort sizes.

\section*{Acknowledgments}

CK developed the proposed model and implemented the ML building blocks used in the study; CK and SM ran the ML analyses. GH performed the neural data analysis. MC developed and ran the FAU pipeline. SM, MB, ASW and SS designed the common IAT task template. MB and SM defined the E-IAT variant, SS and SM defined the PSY-IAT variant, and SM implemented the tasks. ASW, MB, and SS conceptualized both studies. MA oversaw development of the FAU pipeline and contributed to the conceptualization of the study. ANM was involved in data collection and performed supplementary analyses. All authors contributed to the writing and editing of the manuscript.

This project was sponsored by the Defense Advanced Research Projects Agency (DARPA) under Cooperative Agreement No.\ N660012324016. The content of the information does not necessarily reflect the position or the policy of the Government, and no official endorsement should be inferred.

\section*{Disclosures}

ASW has multiple patents in the area of psychiatric biomarkers generally; none of these is licensed to any commercial entity. CK, SM, GH, and TM are employees of Intheon but have no financial stake in the outcome of this study; MC and MA are employees of Deliberate AI but have no financial stake in the outcome of this study. No authors declare a conflict of interest.
\section*{Code and Data Availability}

The authors will make the analysis pipeline and models available upon publication of the article. The NeuroPype software (\texttt{https://neuropype.io}) is freely available for academic use. Due to patient privacy considerations, some components of the dataset analyzed here can not be shared freely at this time, but the authors commit to reasonably supporting reproducibility efforts upon request.

\appendix

\section{Detailed E-IAT and PSY-IAT Task Design} \label{app:taskdetails}

Our stimulus set consisted of 30 unique words for the E-IAT (28 for the PSY-IAT) drawn from four categories: the two concept categories (i.e., trapped, free in the E-IAT and schizophrenia / not schizophrenia in the PSY-IAT), and two attribute categories (me, not-me in both tasks). The concept-attribute pairs (trial categories) were ``trapped/me'' and ``free/me'' in the E-IAT task, and ``schizophrenia/me'' and ``NOT schizophrenia/me'' in the PSY-IAT. Note that while the original PSY-IAT \cite{kirschenbaum2022validation} used the term ``psychosis'' in their category labels, we instead use ``schizophrenia'' here because we felt that the general public has a more precise understanding of this term.

Each E-IAT or PSY-IAT block showed all stimuli once. Order was randomized per block. The trial category (concept-attribute pair) alternated across blocks (e.g., block 1 showing ``trapped/me'', block 2 showing ``free/me'', etc.) with the starting trial category balanced across subjects. Therefore, each block containing a trial category considered to be congruent or in agreement with the subject type (i.e., ``schizophrenia/me'' for a psychosis patient) is followed by a block with a trial category that was incongruent, or in conflict with, the same subject type (i.e., ``NOT schizophrenia/me'' for a psychosis patient). The same pattern is continued across all blocks. The 12 blocks were preceded by a short practice block of 10 trials with randomly selected stimuli, at least one from each concept and attribute.

Subjects were asked to press the left-arrow key if the stimulus word was associated with either the concept or attribute, and press the right-arrow key if the word was not associated with the trial category (either the concept or attribute). Since the attribute was always ``me'', only the concept (i.e., ``trapped'' or ``free'') alternated between blocks. While the category from which the stimuli were drawn varied within a block (since all stimuli from all categories were shown in each block), the trial category, that is, the concept-attribute pair, was constant throughout each block. Therefore, when we refer to a block of trials as being ``congruent'' or ``incongruent'', this refers to the concept-attribute pair, and specifically the concept (since the attribute was always ``me''), being in agreement with or conflicting with the subject's characteristics, not the stimulus word. Both the response key assignment and the starting block category were counterbalanced across subjects.

The stimulus (and trial category) was presented until either response button was pressed or 2000 ms elapsed, resulting in a variable trial length with a maximum of 2 seconds. The inter-trial interval (ITI) was 500 ms with 50 ms of jitter.

\section{Baseline Methods Battery Implementation} \label{app:refmethods}

To quantify the performance of the proposed method relative to reference baselines, we set out to replicate a number of standard machine-learning methods that could be applied across one or more of the relevant modalities, with a focus on well-understood and robustly implemented (``battle-tested'') approaches. Since it has rarely been attempted so far to infer psychometric variables in IATs from modalities other than reaction times, we set up a framework that would allow us to adapt existing single-trial methods to the multi-trial framework while keeping the method's mathematical formulation intact.

The approach, which we call here the ``recoding trick'', can be shown to be formally equivalent to the learning and prediction strategy of the Bayesian method (minus the Bayesian parameter modeling), and proceeds as follows: we observe that imposing the ``mirror'' constraint $W_t^C = -W_t^I$ can alternatively be realized by holding the weight matrix fixed across trials and flipping the class label on the negative-associated trials only, i.e., $y_t^{I\prime} = 1 - y_t^I$. We then employ an off-the-shelf ML model (which generally learns a single weight vector) for training on single-trial data, analogously to the Bayesian method, although as a result, we no longer \textit{neither train on nor predict} the participant labels. To see that this yields a meaningful labeling scheme nevertheless, we can inspect the resulting class labels as per table~\ref{tab:A1}. Here we sort the two classes of stimuli (positive self-associated, i.e., ``entrapped/me'' or ``psychosis/me'' and negative-associated, i.e., ``not entrapped/me'' or ``not psychosis/me'') into a $2 \times 2$ matrix by participant-level association (i.e., ground truth at training time) and trial-level association (fixed by exp.\ design). That is, the original label remains the same on the (+) Stim.\ Assoc.\ column and becomes flipped on the ($-$) Stim.\ Assoc.\ column. In each cell we note the original (participant-derived label) and the new (conditionally flipped) label, along with an annotation where we denote whether the \textit{a priori} trial label is congruent with the given participant label or not (note this relationship is between the participant labeling and the stimulus type and does not change as a result of the flipping operation).

\begin{table}[ht]
\centering
\caption{Basic operation of the ``recoding'' trick. The label gets flipped on the right-hand column and becomes equivalent to the (crossed) participant/trial congruency factor.}
\label{tab:A1}
\begin{tabular}{lcc}
\hline
orig $y$ $\to$ new $y'$ (C/I) & (+) Stim.\ Assoc. & ($-$) Stim.\ Assoc. \\
\hline
(+) Participant ($y=1$) & $1 \to 1$ (C) & $1 \to 0$ (I) \\
($-$) Participant ($y=0$) & $0 \to 0$ (I) & $0 \to 1$ (C) \\
\hline
\end{tabular}
\end{table}

The purpose of the exercise is to make it clear that the transformed label becomes equivalent to the ``congruency'' factor after the label flipping (I=0, C=1). As a result, when using the recoding trick we label trials effectively based on whether they are congruent with the participant ground truth or not, and the model is consequently trained as a congruency detector.

The resulting model will however now also predict congruency at a single-trial level, rather than the desired estimated participant self-association. This can however be rectified by performing the inverse mapping as a post-processing operation on the predicted probabilities (or class labels) generated by the ML model; that is, we reverse ($1-p$) the predicted probability $P(\mathrm{congruent})$ on the trials with negative label-stim associations only. Since this is formally the reverse operation of the label mapping, the resulting predicted label is transformed back to the estimated participant-level association, i.e., $P(\mathrm{positive})$. This can also be seen by working backwards through the above table from the predicted congruency label. After having reversed the prediction, we obtain, for each session, a single-trial prediction, $P(\mathrm{positive})$. We then mimic the Bayesian recipe for integrating these probabilities across trials into a joint probability. We do this by \textit{averaging} the logit scores underlying the predicted probabilities (which can be obtained from the ML method's generated probability estimate by means of the logit function or directly from the underlying decision score), and mapping the averaged score back to a probability by means of the logistic sigmoid function. We note that this complexity could be avoided if the methods were instead reformulated to have a per-trial sign variable built in, as does the Bayesian method, but the chosen approach serves to be able to reuse existing off-the-shelf ML methods and their existing proven implementations (e.g., scikit-learn, EEGNet). We also tested a configuration where we skip the recoding trick and both train and test directly on the participant labels, which, notably, means that all trials in a given session carry the same label. This is the sLDA (direct) configuration in Table~\ref{tab:R2}.

The above approach suffers from the same confidence calibration mismatch of the Bayesian model, and this can in turn be rectified using a Platt scaling approach as discussed in the main text; we are not exploring or evaluating this here, since these methods merely serve as reference baselines, and the scaling affects none of the main metrics that we rely on for comparisons (AUC, Sensitivity, Specificity).

With these preliminaries out of the way, we now turn to the chosen baseline methods. For EEG, we focus on two linear methods and one non-linear (deep learning) method. As the linear methods we replicate (1) the well-known shrinkage Linear Discriminant Analysis (sLDA) method described in \cite{blankertz2011single}, which is considered a simple but effective gold standard for linear EEG phenomena (event-related potentials) that enjoys built-in regularization, and (2) a simple $l_2$-regularized logistic regression (L2LR in Table~\ref{tab:R2}) where the regularization parameter was estimated in a nested cross-validation (using the standard scikit-learn implementation).

Of these, sLDA requires the practitioner to pre-specify a set of time-window features, which represent task-specific prior knowledge that may in practice be difficult or impossible to choose optimally for a task such as the IAT by means other than trial and error. This is not ideal for a number of reasons, including inflated performance when hill-climbing on the same dataset also used for testing. We use here a set of windows that is based on visual inspection of the neural phenomena (which nominally is a whole-data statistic and thus in a strict sense fundamentally ``tainted''), but was not subjected to trial-and-error tuning or other forms of parameter search. We use the following 5 carefully chosen time windows here: $-0.05$--$0.05$; $0.05$--$0.2$; $0.2$--$0.3$; $0.3$--$0.4$; $0.4$--$0.5$ (seconds relative to stimulus onset). In brief summary, the per-channel EEG is averaged in each of these windows to yield a feature, and features are concatenated across channels. The shrinkage parameter ($\lambda$) is automatically inferred on the respective training data using the Ledoit-Wolf method \cite{ledoit2004wellconditioned}. We use a separately validated scikit-learn implementation of sLDA.

As a representative deep learning approach we use the well-known EEGNet approach first proposed in \cite{lawhern2018eegnet}; EEGNet is known to be applicable to both event-related potentials and oscillatory EEG phenomena, or combinations thereof. EEGNet has a number of parameters, although the idea is that a similar set of parameters are intended to work well across multiple tasks. Here we use the same parameters as used in the original paper on ERPs (F1=2, F2=2, kernelSize=25, D=2), along with an AdamW optimizer (LR=0.01, $\beta_1$=0.9, $\beta_2$=0.999, weight decay=0.5), batch size=32, at most 50 epochs with early stopping, a 25\% validation split (retaining a contiguous set of 25\% of training trials after sorting by participant, thus largely splitting by participants for the purpose of early stopping) and a conventional sigmoid binary cross-entropy loss. Some alternative parameter settings for F2, kernelSize, learning rate, and the batch size were explored but did not yield a meaningful improvement. This was implemented using libraries in the JAX ecosystem \cite{deepmind2020jax}, specifically Haiku and Optax, using NeuroPype as the implementation framework.

To use the above methods on multi-modal data also, we applied the following straightforward extensions, where we treat time series from the eye tracker and facial action system as relatively slow-varying signals with event-locked dynamics on a time scale roughly comparable to EEG event-related potentials.

For the logistic regression we vectorized the extracted segment for a given trial for each modality and concatenated the resulting feature vectors across modalities, while for sLDA, we separately extract time-slice average features in each modality (using potentially modality-specific time windows) and then concatenate these features before the LDA stage. Since in non-EEG modalities a restriction to a ``preconscious'' time window (0--400\,ms) may not be as meaningful as in EEG, we test here both the same window set as used for EEG, and a second set of windows that is a superset which additionally extends from $-200$\,ms to $+1000$\,ms relative to the stimulus ($-0.2$--$-0.05$; $0.0$--$0.05$; $0.05$--$0.2$; $0.2$--$0.3$; $0.3$--$0.4$; $0.4$--$0.5$; $0.5$--$0.7$; $0.7$--$1.0$).

For EEGNet, we duplicate the ``front-end'' layers up to but excluding the final classifier layer for each modality, and then fuse representations before that classifier layer, representing a ``late fusion'' approach, keeping all other details the same.

\section{Estimating Training-Data Effective Sample Size} \label{app:samplesize}

We attempted to quantify the effective sample size within each session ($N_{\mathrm{eff}}$) after accounting for inter-trial correlations. To this end, we estimated the design effect (DEFF, a sample-size inflation factor; \cite{kish1965survey}) of one of our IAT datasets using Kish's approximation. This formulation is univariate, and when working with multivariate data, a conservative approach is to choose the worst-case DEFF across all variables. However, since in our case we have a large number of highly \textit{correlated} variables (generally over time and, in case of EEG, also over space/channels), we instead calculate the DEFF for each of the top-5 principal components of our data and retain the worst-case DEFF estimate among the five constructed latent variables. Using larger numbers of PCs (we tested up to 10) did not change the result appreciably. This identifies the worst-case latent dimension, and yields a value on the order of DEFF $\approx 55$. This in turn suggests that the effective number of independent observations (trials) per session may be as low as the raw within-session trial count divided by this correction factor, that is, $270/55 \approx 4.9$ samples. Thus, the total evidence contained in a given session (average single-trial evidence times 4.9) may be closer to the average single-trial evidence, than to a naive sum of evidence across all trials (single-trial evidence times 270 when assuming full independence).

A consequence is that, for a 50-participant/session dataset, we may conservatively expect to have an effective total sample size (across all trials in the training data) on the order of around $50 \times 4.9 \approx 245$ samples. When this is put in relation to the effective model complexity (effective degrees of freedom $df$), it becomes immediately clear that a 500-parameter model would be considered severely under-determined in standard statistical practice (observations per variable $< 1$), unless highly parameter-efficient priors were employed. This line of reasoning informed our reliance on sparse (and possibly smooth) priors, which can be shown to have much lower effective degrees of freedom than the number of variables in the model; see also \cite{zou2007degrees}, \cite{yuan2016degrees}, and \cite{mazumder2020computing} for some attempts to quantify this in models related (but not identical) to ours.

\section{Supplementary Figures} \label{app:suppfigs}

\begin{figure}[htbp]
\centering
($a$)\\[2pt]
\includegraphics[width=\figwidth]{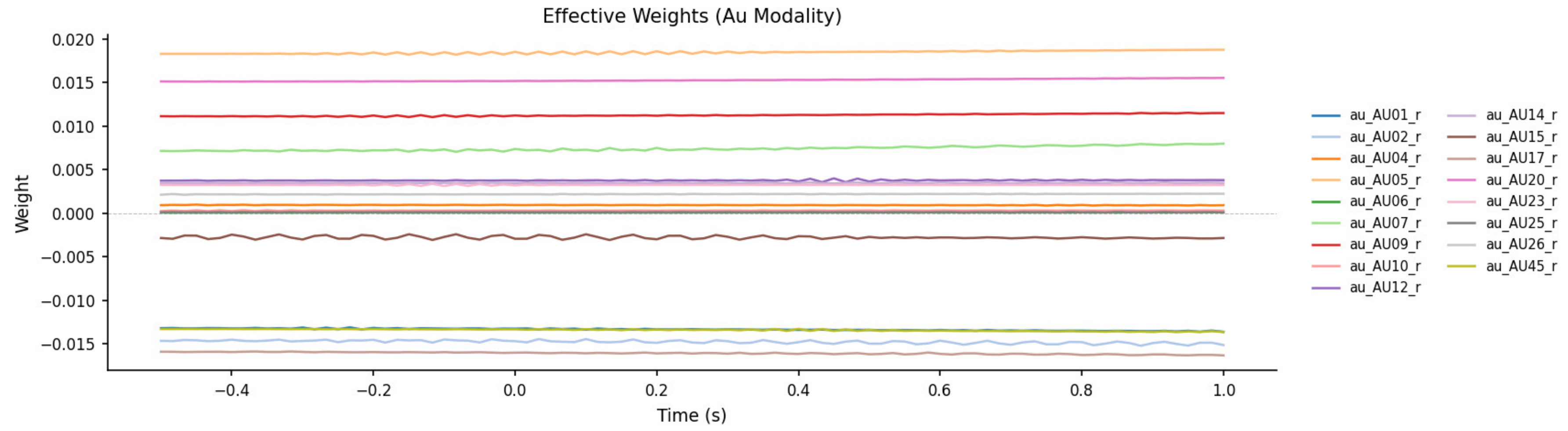}

\vspace{6pt}

($b$)\\[2pt]
\includegraphics[width=\figwidth]{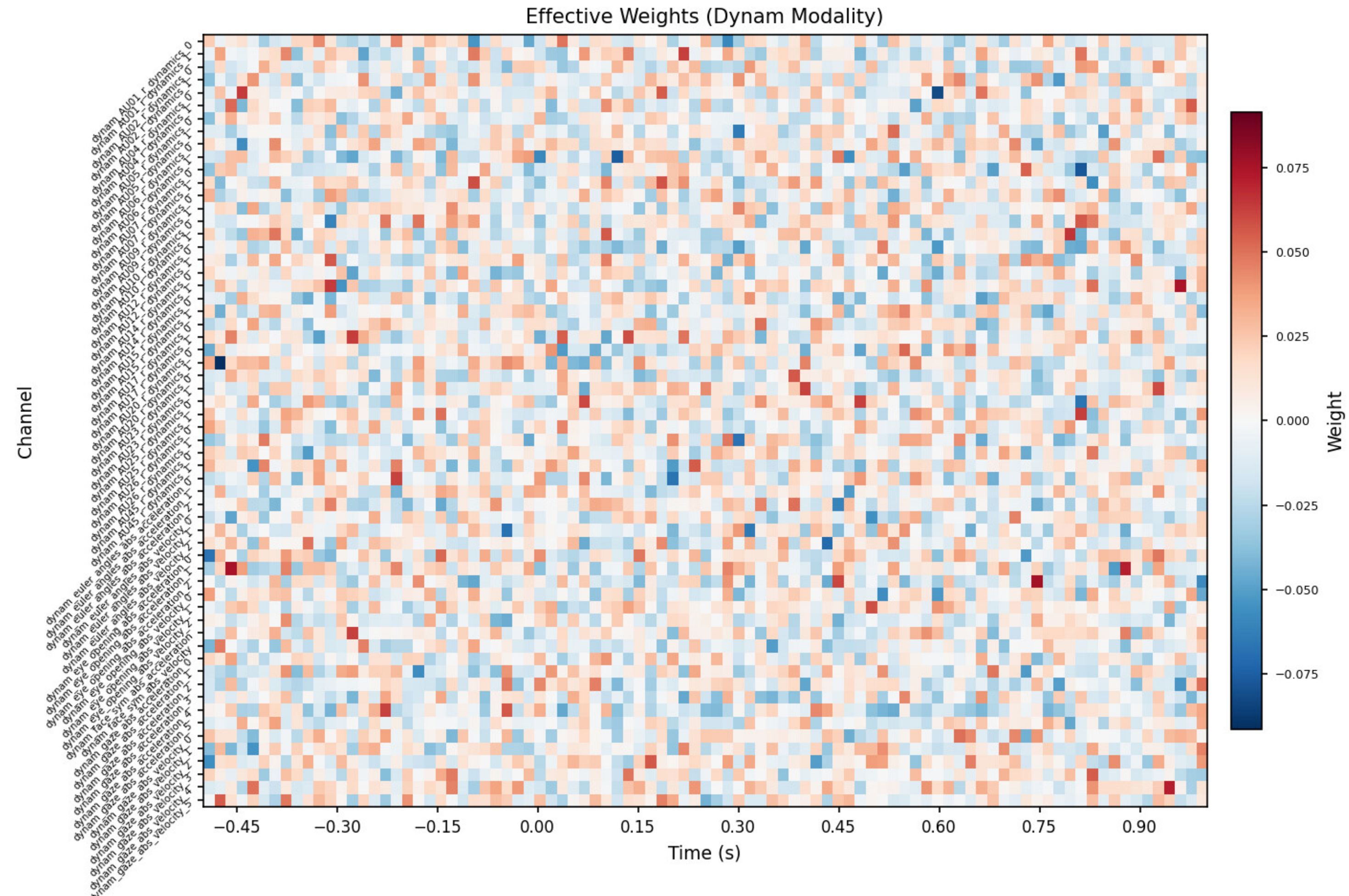}
\caption{Weights for the PSY-IAT model using Dynam+AU modalities; ($a$) AU weights over time (note the high stationarity), ($b$) matrix of Dynam weights (noisy).}
\label{fig:weights-psyiat}
\end{figure}

\bibliography{refs}

\end{document}